\documentclass{article}

\PassOptionsToPackage{numbers,compress}{natbib}
\usepackage[dblblindworkshop, final]{neurips_2026}

\usepackage[utf8]{inputenc}
\usepackage[T1]{fontenc}
\usepackage{hyperref}
\usepackage{url}
\usepackage{booktabs}
\usepackage{array}
\usepackage{amsfonts}
\usepackage{nicefrac}
\usepackage{microtype}
\usepackage{xcolor}
\usepackage{graphicx}
\usepackage{caption}
\usepackage{subcaption}
\usepackage{amsmath}
\usepackage{amssymb}
\usepackage{multirow}
\usepackage{enumitem}
\usepackage{threeparttable}
\usepackage{float}
\usepackage{placeins}
\usepackage{pdflscape}

\title{RankShift: Evaluating Ranking Instability Across Benchmark Dimensions in Document Parsing}
\workshoptitle{TAE (Trust-AI-Eval): Can We Trust AI Evaluation?}
\makeatletter
\renewcommand{\@trackname}{\@neuripsordinal\ Conference on Neural Information Processing Systems (NeurIPS \@neuripsyear). Workshop: \@workshoptitle}
\makeatother

\author{%
  Halimat Afolabi \\
  Stratified Precision \\
  \texttt{halimat@stratifiedprecision.com}
  \And
  Daeyoung Lee \\
  University College London \\
  \texttt{dae.lee.24@ucl.ac.uk}
}

\begin{document}

\maketitle

\begin{abstract}
Document parsing benchmarks are often treated as stable measures of model quality, yet rankings may depend substantially on benchmark design choices. In this work, we present RankShift, a systematic study comparing OCR and vision-language models (VLMs) under controlled variation in dataset composition, matching strategy, and scoring methodology. Using OmniDocBench as a case study, we evaluate the ranking stability of 16 models across 3 dataset configurations, 6 matching strategies, and 3 scoring metrics. We quantify rank stability using pairwise rank correlation, top-$k$ churn, and pairwise reversal rate. Our results show that the interaction between matching strategy and ground truth format is a substantial but understudied driver of instability: scoring against OmniDocBench’s JSON annotations versus its Markdown rendering yields a mean Kendall’s $\tau \approx 0.33$ under region-level matching strategies, with individual pairs as low as $\tau = 0.23$, compared with $\tau = 0.77$ when each page is scored as a single concatenated string. Dataset composition produces comparable instability across page-type strata (mean $\tau = 0.48$) and layout-type strata (mean $\tau = 0.49$), while the cross-dataset shift from OmniDocBench to Real5's Scanning reconstruction of the same pages is more stable ($\tau = 0.80$). Metric choice is the most stable dimension, with mean $\tau \approx 0.78$ across NED, BLEU, and METEOR. These findings indicate that leaderboard position is not solely a property of model  capability, motivating the need for leaderboards  that not only report headline scores, but also the robustness of rankings across reasonable design choices.
\end{abstract}

\section{Introduction}

Benchmarks are essential tools for tracking progress in machine learning, yet they are often treated as static and objective arbiters of model capability~\cite{cais2026hle,qian2026benchmark2,eriksson2025trust}. In practice, benchmarks encode a range of design choices, including dataset composition, pre- and post-processing steps, and scoring methodology. The effects of these decisions on model rankings are rarely made explicit and remain largely understudied in document parsing. As benchmarks become increasingly saturated and susceptible to overfitting, these choices take on greater significance: small differences in evaluation design can determine a model's overall ranking, independent of genuine capability differences.

Document parsing, the task of converting unstructured document images into structured, machine-readable output, is particularly challenging to evaluate. Models vary substantially in architecture, output format, and how they encode complex elements such as tables and formulas. These differences mean that the way a benchmark evaluation pipeline segments predictions, matches them to ground truth, and scores them can systematically advantage or disadvantage entire classes of systems.

Data distribution shift is a well recognised challenge for benchmark validity~\cite{taori2020measuring}, and scoring metrics have also been widely scrutinised~\cite{mathur2020tangled}, though their effect on model rankings remains understudied in document parsing, where output heterogeneity makes such choices especially consequential. Alignment strategies, the procedures by which model predictions are matched to ground truth before scoring, are almost never compared, despite being the axis on which document parsing differs most from other evaluation settings. 

OmniDocBench~\cite{ouyang2024omnidocbench} has become the de facto benchmark for comparing 
document parsing systems, yet its own maintainers have identified matching-strategy bias, the 
penalising of correct content that is segmented differently from the annotation, as an important factor limiting its effectiveness~\cite{mineru25}. Version 1.6 introduces a redesigned matching 
procedure in response. However, neither the bias nor the fix has been quantified, and their 
effect on model rankings remains unknown.

In this work, we present \textbf{RankShift}, a systematic study of ranking instability in document parsing evaluation. Using the OmniDocBench v1.5 dataset with the v1.6 evaluation protocol as a case study, we evaluate 16 OCR and vision-language models under controlled variation across three benchmark dimensions: alignment strategy, dataset composition, and scoring metric. For alignment, we compare six strategies formed by crossing three matching methods with two ground-truth formats: end-to-end (E2E), which scores predictions against OmniDocBench's JSON annotations, and markdown-to-markdown (M2M), which scores them against the Markdown rendering of those annotations. For dataset composition, we compare rankings across the full benchmark, its nine page-type strata, five layout-type strata, and a scanned reconstruction of the same pages (Real5~\cite{zhou2025real5}). For scoring, we compare three string based evaluation metrics: NED, BLEU, and METEOR.

\begin{figure*}[t]
\centering
\includegraphics[width=\textwidth]{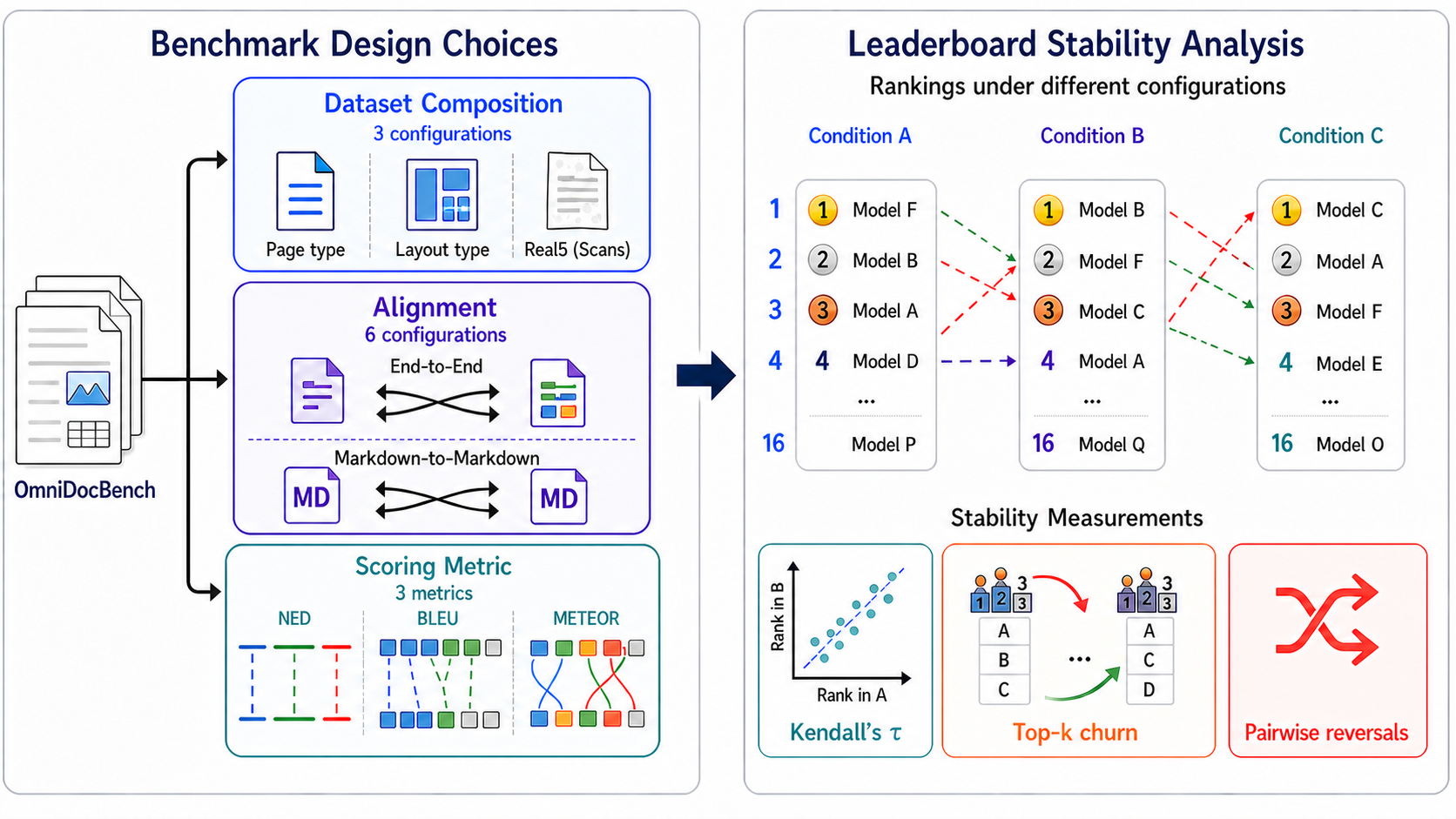}
\caption{Evaluation dataset, alignment strategy and scoring metric are 
varied independently, and the resulting leaderboards compared using Kendall's $\tau$, top-$k$ 
churn, and pairwise reversals.}
\label{fig:rankshift-overview}
\end{figure*}

Figure~\ref{fig:rankshift-overview} summarizes how these benchmark design choices produce alternative model rankings and how RankShift measures their stability.

Our contributions are as follows:

\begin{itemize}
    \item We present the first study of ranking instability in document parsing, quantifying how model rankings shift under reasonable variation in evaluation design.
    \item We identify the interaction between ground-truth format and matching method as an understudied source of ranking instability that, under segment-level matching, reverses nearly a third of all model pairs --- rivalling dataset composition and substantially exceeding metric choice.
    \item We propose a diagnostic procedure,  based on a model's profile of scores across alignment strategies, in which recurrent disagreement patterns suggest candidate failure modes such as hallucinated content or segmentation mismatch.
\end{itemize}
\section{Related Work}

\subsection{Document Parsing Benchmarks}

Appraising a parsed page fairly is difficult because  multiple valid representations exist for the same underlying content. This has led to substantial divergence in how 
leading document parsing benchmarks score parsed outputs. olmOCR-Bench~\cite{olmocr} and 
ParseBench~\cite{parsebench} replace text-similarity scoring with binary unit testing, arguing that the former penalise trivial formatting differences~\cite{wang2025imageovertext} 
while missing failures that matter downstream~\cite{parsebench}. Each test asserts a single property such as required text, reading order or table structure. In contrast, 
DOCR-Inspector~\cite{zhang2025docrinspector} uses a vision-language model (VLM) as a judge to 
assess parsing quality across 28 error types. These approaches avoid whole-page matching of parsed 
outputs to reference annotations, and so avoid the difficulties of global alignment, while making each assertion or error label 
individually interpretable.

While these approaches mitigate the problems of global alignment, they introduce a closed-world assumption: text-similarity scoring accounts for every divergence between a prediction and its reference, whereas assertion- and judge-based approaches can only identify the failure modes they were 
designed to detect, so their coverage is fixed by design rather than by the output itself. DOCR-Inspector's error taxonomy derives from manual analysis of a sample of documents and, though 
broad, is unlikely to capture all error types.  Additionally multi-error examples form only 16\% of its training data, synthesised from two to four 
non-interfering error types, whereas real-world errors are both more common and more likely to interact. 
The unit-test-based benchmarks inherit a similar selection bias: a test exists only where the 
generating model could read the content, so pages that defeat the generator yield fewer and easier 
assertions. OmniDocBench~\cite{ouyang2024omnidocbench}, by contrast, appraises output by matching 
full-page predictions to annotated ground truth and scoring the matched spans with 
text-similarity metrics. It remains the gold-standard benchmark against which most document 
parsing systems are compared.

\subsection{OmniDocBench and its Ecosystem}

OmniDocBench~\cite{ouyang2024omnidocbench} is unique in the range of evaluation configurations it exposes: 
three matching strategies, two ground-truth formats, a choice of metrics, and attribute-level 
stratification. Its documentation justifies its default evaluation method only on the grounds 
that the JSON format preserves the category and attribute annotations that enable category-level 
filtering and attribute-stratified results, without 
discussing how any of these choices affects scores or rankings. The maintainers have since identified matching-strategy bias as a limitation, whereby matching predictions to annotations one-to-one penalises systems whose 
segment boundaries differ from the annotations', even where the parsed content is correct. 
To address this, version 1.6 introduces Multi-Granularity Adaptive Matching (MGAM), which holds 
the ground truth fixed and searches for an optimal segmentation granularity on the prediction 
side~\cite{mineru25}. The redesign is defined rather than evaluated, however: no comparison of 
scores or model rankings under the different matching strategies is reported, and it remains 
unclear which output characteristics each strategy systematically rewards or penalises, and why.

Several benchmarks inherit OmniDocBench's  evaluation methodology including InduOCRBench~\cite{induocrbench2026}, MDPBench~\cite{mdpbench2026}, MPDocBench-Parse~\cite{mpdocbenchparse2026} and Dr. DocBench~\cite{drdocbench2026}. Some retain both ground-truth formats without 
discussing the trade-off, while others adopt one silently. None appear to weigh the consequences of the alignment or metric choice, a decision that grows more consequential as the ecosystem widens.

\subsection{Benchmark Sensitivity and Ranking Instability}

Ranking instability under evaluation design variation is a recognised but understudied problem, 
though a growing body of work has begun to address it~\cite{eriksson2025trust}. \citet{post2018call} and
\citet{marie2021scientific} highlight the difficulty of comparing even nominally identical scores 
such as BLEU across the literature, since implementation choices such as reference text 
preprocessing leave them effectively non-standardised. Similarly, 
\citet{alzahrani-etal-2024-benchmarks} showed that minor perturbations in popular multiple-choice question
benchmarks, including changing the order of answers or the means of selecting them, can 
substantially alter LLM leaderboard rankings.

Benchmarks themselves have received similar scrutiny. The Benchmark 
Lottery~\cite{dehghani2021benchmarklottery} argues that rankings are fragile to benchmark task 
selection biases, which can unfairly favour models whose strengths align with the chosen tasks, at the expense of those that perform well on others. Benchmark$^2$~\cite{qian2026benchmark2} attempts to quantify benchmark 
quality directly by evaluating benchmarks along three axes: cross-benchmark ranking consistency (how well the rankings a benchmark produces correlate with 
those of peer benchmarks), discriminability (how well it distinguishes between models), and 
capability alignment deviation (identifying problematic instances where strong models do worse 
than weaker models from the same family), and found significant variance in quality among widely 
used benchmarks. Attention has since turned to individual fields: SafetyRepro~\cite{li2026safetyrepro} found that 
on every AI safety alignment benchmark tested, configuration choice alone can flip which model is 
deemed safer, and \citet{kendiukhov2026grn} found pairwise reversal rates from 0.0\% to 32.1\% 
across four evaluation protocol axes in gene regulatory network benchmarking.

Together, these studies establish that benchmark rankings are fragile to evaluation design. However, 
its impact in document parsing, which carries the distinctive difficulty of aligning model outputs 
to reference text, remains underexplored.

\section{Methodology}

\subsection{Datasets}

We evaluate all models on two datasets: the full OmniDocBench v1.5 benchmark and the Real5-OmniDocBench Scanning subset, which together support our three dataset configuration experiments.

\paragraph{OmniDocBench v1.5.}
OmniDocBench~\cite{ouyang2024omnidocbench} is the primary benchmark used throughout this work. Version 1.5 comprises 1,355 bilingual PDF pages spanning nine document types; academic papers, textbooks, financial reports, research reports, slides, books, newspapers, exam papers, and handwritten notes, across five layout types. Pages are annotated at 19 block-level categories and 15 attribute labels, enabling end-to-end, module-level, and attribute-stratified evaluation.

\paragraph{Real5-OmniDocBench (Scanning).}
Real5-OmniDocBench~\cite{zhou2025real5} performs a full-scale, one-to-one physical reconstruction of all 1,355 OmniDocBench v1.5 pages across five real-world degradation conditions: Scanning, Warping, Screen-Photography, Illumination, and Skew. Only the Scanning condition, which represents the most common real-world document capture scenario, is used in the study. The one-to-one correspondence with OmniDocBench v1.5 is critical to our experimental design: because every scanned page shares its ground truth annotation with the digital counterpart, differences in model rankings between the two datasets can therefore be attributed to capture condition alone.

\subsection{Models}

We evaluate 16 models spanning specialist document and general-purpose VLMs (Table~\ref{tab:models}).

\begin{table}[t]
\centering
\caption{Models evaluated in RankShift. Type: P = Pipeline, S-VLM = Specialist VLM, G-VLM = General-Purpose VLM. Output format indicates the primary structured output produced.}
\label{tab:models}
\resizebox{\columnwidth}{!}{%
\begin{tabular}{llcccc}
\toprule
\textbf{Model} & \textbf{Type} & \textbf{Params} & \textbf{Open} & \textbf{Doc-Specialised} & \textbf{Output} \\
\midrule
Docling            & P      & --    & $\checkmark$ & $\checkmark$ & Markdown \\
MinerU2.5          & S-VLM      & 1.2B  & $\checkmark$ & $\checkmark$ & Markdown \\
Chandra 2          & S-VLM  & 5B    & $\checkmark$ & $\checkmark$ & Markdown \\
DeepSeek-OCR 2     & S-VLM  & 3B    & $\checkmark$ & $\checkmark$ & Markdown \\
Dolphin 1.5        & S-VLM  & 0.5B  & $\checkmark$ & $\checkmark$ & Markdown \\
dots.ocr           & S-VLM  & 1.7B  & $\checkmark$ & $\checkmark$ & Markdown \\
GLMOCR             & S-VLM  & 0.9B  & $\checkmark$ & $\checkmark$ & Markdown \\
GOT-OCR2.0         & S-VLM  & 580M  & $\checkmark$ & $\checkmark$ & MD/LaTeX/HTML \\
Hunyuan OCR        & S-VLM  & 1B    & $\checkmark$ & $\checkmark$ & Markdown \\
MonkeyOCR-pro-3B   & S-VLM  & 3B    & $\checkmark$ & $\checkmark$ & Markdown \\
PaddleOCR-VL-1.5   & S-VLM  & 0.9B  & $\checkmark$ & $\checkmark$ & Markdown \\
RolmOCR            & S-VLM  & 8B\textsuperscript{\dag} & $\checkmark$ & $\checkmark$ & Markdown \\
Youtu-Parsing      & S-VLM  & 2.5B  & $\checkmark$ & $\checkmark$ & Markdown \\
Gemini 2.5 Pro     & G-VLM  & --    & $\times$ & $\times$ & Markdown \\
Gemini 3.1 Flash   & G-VLM  & --    & $\times$ & $\times$ & Markdown \\
GPT-5.5            & G-VLM  & --    & $\times$ & $\times$ & Markdown \\
\bottomrule
\multicolumn{6}{l}{\textsuperscript{\dag}RolmOCR is a fine-tuned Qwen2.5-VL-7B-Instruct model.}
\end{tabular}%
}
\end{table}

\textbf{Pipeline-based parsers.} Docling~\cite{docling2024} chains layout analysis, table recognition, and reading order prediction to produce structured markdown.

\textbf{Specialist document VLMs.} This group spans a range of scales and architectures: MinerU2.5~\cite{niu2025mineru25} (1.2B), GOT-OCR2~\cite{wei2024gotocr} (580M), GLM-OCR~\cite{glmocr2026} (0.9B), PaddleOCR-VL-1.5~\cite{cui2025paddleocrvl} (0.9B), Dolphin~1.5~\cite{feng2025dolphin} (0.5B), MonkeyOCR-Pro-3B~\cite{li2025monkeyocr} (3B), DeepSeek-OCR~2~\cite{deepseekocr2025} (3B), Chandra~2~\cite{chandra2025} (5B), HunyuanOCR~\cite{hunyuan2025ocr} (1B), Youtu-Parsing~\cite{cao2026youtuparsing} (2.5B), dots.ocr~\cite{li2025dotsocr} (1.7B), and RolmOCR~\cite{rolmocr2025} (8B), a fine-tuned Qwen2.5-VL-7B-Instruct variant. All are trained specifically for structured document output and produce markdown.

\textbf{General-Purpose VLMs.} We evaluate three closed-source frontier models via API using 
the inference scripts and prompts provided in the OmniDocBench repository: GPT-5.5~\cite{openai2025}, Gemini 2.5 Pro~\cite{comanici2025gemini}, and Gemini 3.1 Flash~\cite{google2026gemini31flash}. These serve as general-purpose baselines, allowing us to assess whether ranking instability is specific to specialized document parsing models or reflects a property of the evaluation pipeline independent of model class.

A complete summary of model families, parameter scales, open-source availability, and output formats is provided in Table~\ref{tab:models}.

\subsection{Experimental Setup}

\paragraph{Experiment 1 --- Dataset Dimension.}
We examine ranking stability under three types of dataset variation.
First, \emph{page type}: we partition OmniDocBench into its nine page-type strata
(e.g.\ academic papers, newspapers, textbooks) and measure pairwise rank
stability between every pair ($\binom{9}{2} = 36$ comparisons).
Second, \emph{layout}: we partition the same pages into five layout strata
(e.g.\ single-column, double-column, three-column) and repeat the analysis
($\binom{5}{2} = 10$ comparisons).
Third, \emph{capture condition}: we compare rankings on the original digital pages
against the physically re-scanned Real5 pages.
Strata are disjoint, so partition-based rankings are derived from entirely
different pages, whereas the Real5 comparison holds page content fixed and
varies only how the pages were captured.
Table~\ref{tab:dataset-configs} in the appendix lists all configurations.

\paragraph{Experiment 2 --- Alignment Dimension.} 
\label{sec:exp2}
We compare six alignment strategies, formed 
by crossing three matching methods with two ground-truth formats. In the four segment-level conditions, predictions are 
segmented on double line breaks before matching; the methods differ in how those segments are 
then matched to ground-truth segments: \texttt{quick\_match} (adjacency-search-based truncation 
and merging of adjacent predicted segments, with hybrid formula--text matching to reduce 
scoring sensitivity to unicode formula output), \texttt{simple\_match} (one-to-one assignment 
without truncation or merging), and \texttt{no\_split} (whole-page concatenation into a single string, with no block-level alignment or attribute-level scoring). Each method is 
applied under two ground-truth formats: end-to-end (E2E), which scores against OmniDocBench's 
JSON annotations, and markdown-to-markdown (M2M), which scores against their Markdown 
rendering. This yields $\binom{6}{2}=15$ pairwise comparisons.

The \texttt{quick\_match} condition reported across all experiments uses the v1.6 implementation, which incorporates 
MGAM~\cite{mineru25}. The redesign applies only to \texttt{quick\_match}: its prediction-side granularity search 
presupposes that predicted segments can be merged or split to fit the ground truth, which 
\texttt{simple\_match} and \texttt{no\_split} exclude by definition. The \texttt{simple\_match} 
and \texttt{no\_split} procedures are therefore unchanged from v1.5.

\paragraph{Experiment 3 --- Metric Dimension.}
Under a fixed alignment (\texttt{quick\_match} E2E), we compare three scoring metrics:
normalised edit distance (NED, reported as $1 - \text{NED}$), BLEU, and METEOR.
All three are computed on the same aligned text segments, so rank differences reflect
metric sensitivity rather than matching or pipeline choices.

\paragraph{Metric underlying reported rankings.} Although OmniDocBench reports a composite score averaging text edit distance, table TEDS and formula CDM~\cite{wang2025imageovertext}, all rankings analysed in the main text are derived from text edit distance alone, unless stated otherwise. This is because tables and formulas are extracted into separate categories before the text matcher runs~\cite{ouyang2024omnidocbench}: the merging and splitting that distinguishes the alignment conditions acts only on the remaining text paragraphs. Reporting a composite would therefore dilute the manipulation.

\subsection{Alignment Signatures}
\label{sec:signatures}

To characterise \emph{where} alignment strategies disagree, we represent every complete page$\times$model row by a six-character \emph{alignment signature}. Page-level NED scores under the three E2E and three M2M alignments are discretised into zero, low, medium, or high buckets and concatenated in fixed order. Across 20,610 complete rows, this produces 301 observed signatures; the 20 most frequent cover 78.5\% of rows (Figure~\ref{fig:signature-top20-buckets}).

We combine signature frequency, score-cluster shape, corpus-level text statistics, and stratified manual review to associate recurrent E2E--M2M disagreement patterns with candidate failure mechanisms. ~\ref{app:signature-details} provides the complete construction, thresholds, auxiliary labels, sensitivity analysis, and review protocol.

\subsection{Stability Measures}

We measure ranking instability between any two conditions $A$ and $B$ using three
complementary statistics computed on the vector of per-model mean scores.

\textbf{Kendall's $\tau$} measures global rank correlation ($\tau = 1$ indicates identical
ordering; $\tau = 0$ is consistent with statistical independence).

\textbf{Top-$k$ churn rate} quantifies instability at the top of the leaderboard:
\begin{equation}
  \text{churn}(A, B, k) = \frac{k - |\mathcal{T}_k(A) \cap \mathcal{T}_k(B)|}{k},
\end{equation}
where $\mathcal{T}_k(X)$ is the set of top-$k$ models under condition $X$.
A churn rate of 0 indicates the top-$k$ set is unchanged; 1 means complete turnover.

\textbf{Pairwise reversal rate} counts what fraction of model pairs $\{i, j\}$ have their
relative ordering reversed between $A$ and $B$:
\begin{equation}
  \text{rev}(A, B) = \frac{|\{(i,j) : \text{sgn}(s_i^A - s_j^A) \neq \text{sgn}(s_i^B - s_j^B)\}|}{|\{(i,j) : s_i^A \neq s_j^A \wedge s_i^B \neq s_j^B\}|}.
\end{equation}

\subsection{Bootstrap Confidence Intervals}

All three statistics are accompanied by 95\% bootstrap confidence intervals obtained from
$B = 2{,}000$ iterations. In each iteration, pages are resampled with replacement and per-model means are recomputed with multinomial weights. For conditions sharing the same page set (cross-dataset, Experiment~3), a single resample is applied to both conditions simultaneously, preserving page-level pairing. For conditions covering disjoint page sets (strata comparisons), the two strata are resampled
independently, reflecting the true sampling uncertainty for each. For mean $\tau$ across strata, bootstrap resamples are drawn at the model level.
\section{Results}

\begin{table}[t]
\centering
\footnotesize
\setlength{\tabcolsep}{3pt}
\caption{Pairwise Kendall's $\tau$ between model rankings (mean page-level scores). 95\% CIs from bootstrap resampling of models ($n=16$), $B=2000$ replicates.}
\label{tab:kendall_tau}
\begin{tabular}{p{1.4cm} p{6.0cm} p{2.7cm} p{1.8cm}}
\toprule
\textbf{Dimension} & \textbf{Comparison} & \textbf{Kendall's $\tau$ [95\% CI]} & \textbf{Category} \\
\midrule
Alignment & quick\_match vs simple\_match (E2E) & 0.767 [0.575, 0.929] & E2E vs E2E \\
 & quick\_match vs no\_split (E2E) & 0.550 [0.229, 0.821] & E2E vs E2E \\
 & simple\_match vs no\_split (E2E) & 0.583 [0.256, 0.860] & E2E vs E2E \\
 & quick\_match vs no\_split (M2M) & 0.817 [0.593, 0.965] & M2M vs M2M \\
 & simple\_match vs no\_split (M2M) & 0.683 [0.449, 0.892] & M2M vs M2M \\
 & quick\_match vs simple\_match (M2M) & 0.667 [0.368, 0.911] & M2M vs M2M \\
 & quick\_match (E2E) vs quick\_match (M2M) & 0.400 [-0.037, 0.772] & E2E vs M2M \\
 & quick\_match (E2E) vs simple\_match (M2M) & 0.233 [-0.276, 0.699] & E2E vs M2M \\
 & quick\_match (E2E) vs no\_split (M2M) & 0.383 [-0.071, 0.776] & E2E vs M2M \\
 & simple\_match (E2E) vs quick\_match (M2M) & 0.367 [-0.009, 0.682] & E2E vs M2M \\
 & simple\_match (E2E) vs simple\_match (M2M) & 0.300 [-0.207, 0.741] & E2E vs M2M \\
 & simple\_match (E2E) vs no\_split (M2M) & 0.383 [-0.037, 0.736] & E2E vs M2M \\
 & no\_split (E2E) vs quick\_match (M2M) & 0.717 [0.474, 0.912] & E2E vs M2M \\
 & no\_split (E2E) vs simple\_match (M2M) & 0.583 [0.274, 0.852] & E2E vs M2M \\
 & no\_split (E2E) vs no\_split (M2M) & 0.767 [0.504, 0.948] & E2E vs M2M \\
\midrule
Dataset & OmniDocBench v1.5 vs Real5 (E2E quick\_match NED) & 0.800 [0.589, 0.947] & Cross-dataset \\
 & Mean tau across page-type (data\_source) strata, E2E quick\_match & 0.475 [0.279, 0.646] & Intra-dataset \\
 & Mean tau across layout strata, E2E quick\_match & 0.490 [0.247, 0.686] & Intra-dataset \\
\midrule
Metric & NED vs BLEU (E2E quick\_match) & 0.700 [0.463, 0.874] & --- \\
 & NED vs METEOR (E2E quick\_match) & 0.767 [0.486, 0.945] & --- \\
 & BLEU vs METEOR (E2E quick\_match) & 0.867 [0.720, 0.965] & --- \\
\bottomrule
\end{tabular}
\end{table}

\subsection{Ground-Truth Format Substantially Reorders Rankings Under Segment-Level Matching}

\paragraph{Ranking stability within a single ground-truth format, across matching strategies.}
Varying only the matching method within a single ground-truth format produces moderate 
disagreement. Against the JSON annotations, $\tau$ ranges from 0.550 to 0.767, with reversal 
rates of 11.7\%--20.0\%; against the Markdown rendering, agreement is comparable 
($\tau=0.667$--$0.817$), with reversal rates of 9.2\%--16.7\%. None of these six reversal rates 
are significant after FDR correction, indicating that they are consistent with ordinary 
variation within a fixed format (Table~\ref{tab:kendall_tau}; Appendix 
Table~\ref{tab:reversal_significance}).

\paragraph{Ranking stability across ground-truth formats, within a single matching strategy.}
Three pairs hold the matching method fixed and vary only the ground-truth format, isolating its 
effect. Under \texttt{no\_split}, where each page is scored as a single concatenated string, the 
two formats agree closely ($\tau=0.767$, reversal rate 10.0\%, not significant after FDR 
correction). Under the two segment-level methods, agreement collapses: $\tau=0.400$ for 
\texttt{quick\_match} and $\tau=0.300$ for \texttt{simple\_match}. Both formats are derived from the same annotations, and the divergence appears only when 
predictions are matched segment by segment. This is therefore the leaderboard-level consequence of 
the matching-strategy bias identified by the OmniDocBench authors~\cite{niu2025mineru25}.

Across the four pairs in which both conditions match at segment level, mean Kendall's $\tau$ is 
approximately 0.33 and the mean pairwise reversal rate is 31.7\%. The most unstable pair, E2E 
\texttt{quick\_match} versus M2M \texttt{simple\_match}, reverses 35.8\% of model pairs (95\% CI 
$[30.0\%, 39.2\%]$) and has a top-5 churn rate of 0.60, meaning that three of the five models in 
the top set change. Table~\ref{tab:kendall_tau} summarises all pairwise $\tau$ values, and Appendix 
Table~\ref{tab:reversal_significance} reports the corresponding reversal-rate significance tests.

\begin{figure*}[t]
    \centering
    \includegraphics[width=\textwidth]{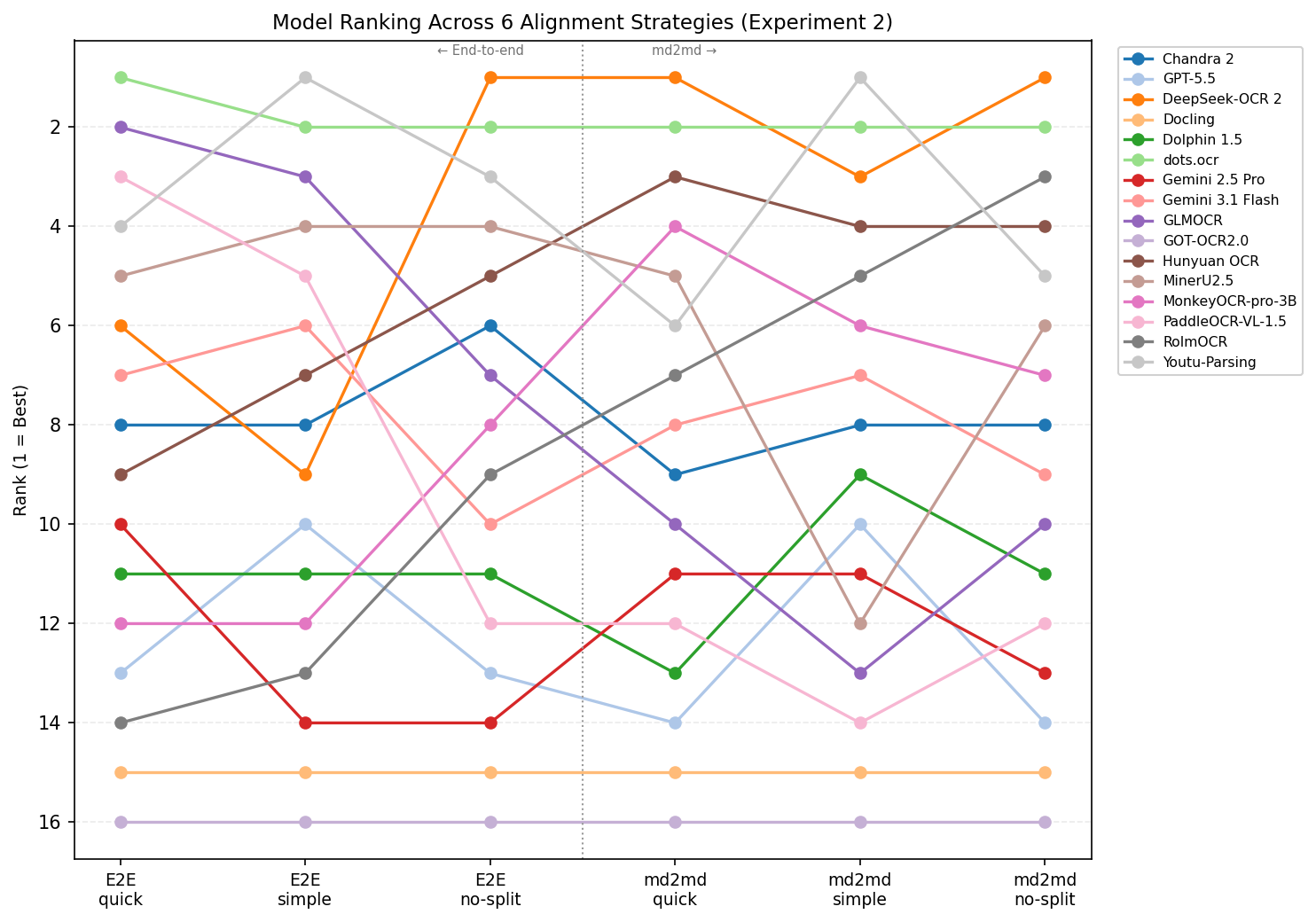}
    \caption{Experiment 2 bump chart showing rank trajectories under alignment variation.}
    \label{fig:bump-exp2-main}
\end{figure*}

\subsection{Diagnostic Protocol}
\label{sec:diagnostic}

The question of which evaluation protocol best agrees with human judgment of parsing quality 
remains open to investigation. Each appears to have distinct strengths and weaknesses that practitioners, benchmark 
creators, and OCR developers should treat explicitly. We therefore frame the E2E--M2M score gap 
as a candidate diagnostic signal, in which the direction and shape of disagreement may indicate 
which failure mode dominates. 

On OmniDocBench 
v1.5, recurrent E2E-favouring families account for approximately 25\% of complete 
page$\times$model rows, compared with approximately 0.8\% for the M2M-favouring families. Our manual review suggests that E2E is comparatively 
lenient on extra or loosely aligned text, whereas M2M can avoid some region-level equation, 
layout, or title-matching failures. The 
asymmetry may therefore reflect corpus composition rather than any general protocol superiority. \ref{app:diagnostic-details} reports the supporting signature clustering, mechanism summaries, manual-review protocol, and diagnostic tables.

\subsection{Dataset Composition Produces Instability Across Page-Type and Layout Strata}

\paragraph{Page-type stratification.}
Splitting OmniDocBench by document type reveals substantially greater instability than the cross-dataset comparison. Under E2E \texttt{quick\_match} NED, the mean agreement across all page-type pairs is $\tau=0.475$, with individual comparisons ranging from $\tau=0.067$ for academic literature versus research report to $\tau=0.717$ for PPT2PDF versus magazine. Academic literature is a systematic outlier, producing the lowest agreement with nearly every other stratum; its near-zero correlation with research reports indicates that these two text-dense document types induce almost unrelated model orderings. This heterogeneity is obscured by aggregate results: dots.ocr ranks first overall but only tenth within the \texttt{academic\_literature} stratum, a nine-position difference.

\paragraph{Layout-type stratification.}
Layout strata exhibit a similar degree of instability (mean $\tau=0.490$, range $[0.267, 0.600]$ under E2E \texttt{quick\_match} NED). The least stable comparison is 1$+$ column versus three-column ($\tau=0.267$), partly reflecting the small size of the three-column stratum (52 pages). The most stable pairs reach only $\tau=0.600$, including 1$+$ column versus double column and \texttt{other\_layout} versus three-column. Stratifying by layout can again reveal strengths the aggregate conceals: Dolphin~1.5 ranks 11th overall 
but fourth on three-column pages.

\paragraph{Cross-dataset (OmniDocBench vs.\ Real5 Scanning).}
For text-based NED, rankings are broadly preserved when moving from digital PDFs to scanned reconstructions ($\tau = 0.800$, 95\% CI $[0.589,\ 0.947]$; pairwise reversal rate = 0.100, CI $[0.067,\ 0.133]$). However, there is still some movement. The top-5 set is not fully preserved (churn rate = 0.20): 
dots.ocr drops out and DeepSeek-OCR~2 enters. Additionally, five models shift by two or more positions: dots.ocr 
falls from 1st to 6th under Real5, MonkeyOCR-pro-3B moves from 12th to 14th, while 
PaddleOCR-VL-1.5, MinerU~2.5 and RolmOCR each improve by two ranks.

\subsection{Metric Choice Is the Most Stable Dimension}

Of the three evaluation dimensions, metric choice produces the least instability. All three pairwise comparisons yield $\tau \geq 0.700$: BLEU versus METEOR achieves the highest agreement ($\tau=0.867$), followed by NED versus METEOR ($\tau=0.767$) and NED versus BLEU ($\tau=0.700$), giving a mean of $\tau\approx0.78$. 

Despite strong overall correlation, individual models still change position under different 
metrics. dots.ocr leads under NED, whereas Youtu-Parsing leads under both BLEU and METEOR, 
rising three places from fourth, suggesting its outputs are rewarded more by n-gram overlap 
than by character-level edit distance. Hunyuan OCR moves from eighth under NED to fifth under 
BLEU, while MinerU 2.5 moves in the opposite direction, from fifth to seventh. Therefore even the most stable dimension tested produces measurable ranking instability, indicating that 
leaderboard positions are better read as approximate than as fixed.

\begin{figure}[t]
    \centering
    \includegraphics[width=\columnwidth]{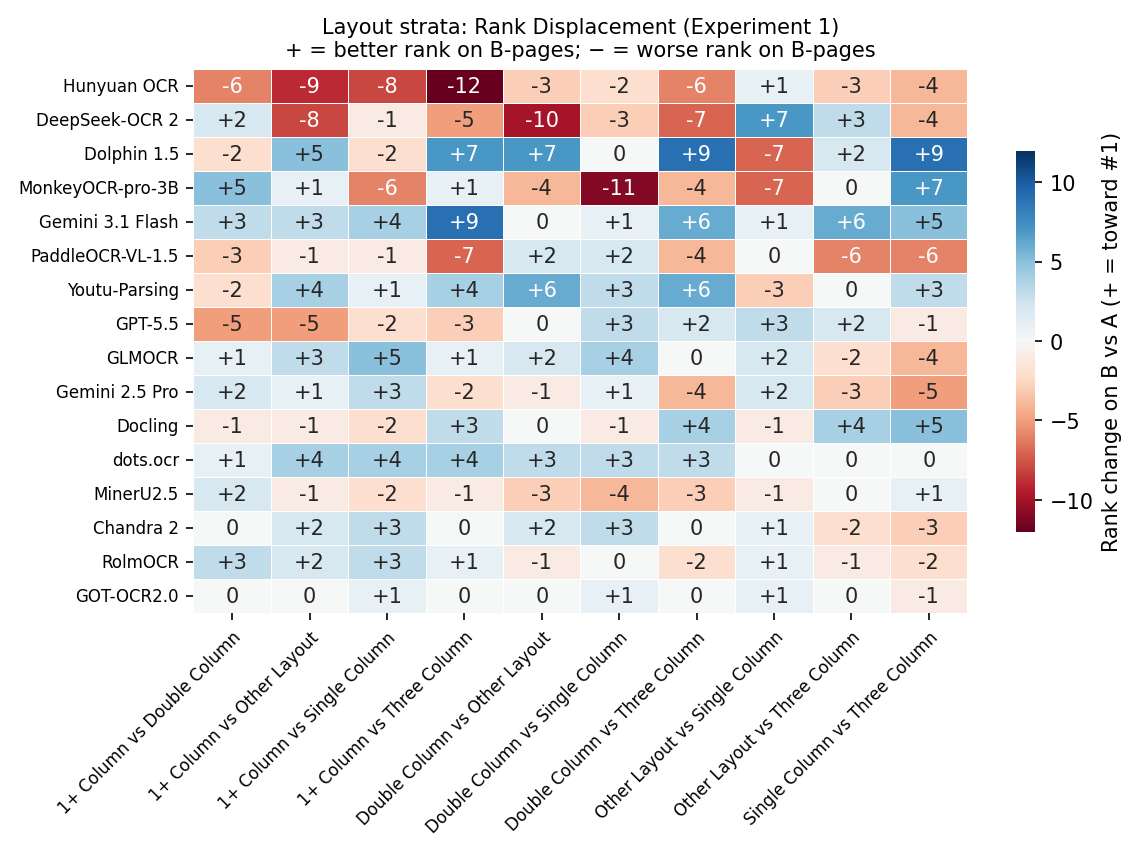}
    \caption{Rank-displacement heatmap for Experiment 1 across layout strata. Each cell shows how many leaderboard positions a model gains when ranked on B-pages instead of A-pages (rank A $-$ rank B). $+12$ (blue) means the model moved 12 places toward \#1; $-3$ (red) means it fell 3 places.}
    \label{fig:rank-displacement-layout-main}
\end{figure}
\section{Discussion}

\subsection{Leaderboard Position Is Highly Conditional}

A model's rank on a document parsing benchmark reflects not only its capability, but also the evaluation configuration under which it was assessed. The same predictions, scored under different alignment strategies, dataset compositions, or metrics, can produce substantially different orderings. A clear illustration of this is the reversal of rankings between GLMOCR and RolmOCR. Under E2E  
\texttt{quick\_match}, GLMOCR ranks 2nd while RolmOCR ranks 14th. Under M2M \texttt{simple\_match}, the positions invert, with GLMOCR falling to rank 13 and RolmOCR rising to rank 5. The direction of this reversal holds across every E2E/M2M pair including the \texttt{no\_split} 
pair, where the two formats on average agree closely (Figure~\ref{fig:appendix-bump-rolmocr-glmocr}), so it reflects the ground-truth format rather than any particular matching method: GLMOCR's output aligns more closely with OmniDocBench's JSON annotations, while RolmOCR's aligns more closely with their Markdown rendering. Applying the diagnostic protocol (\ref{app:diagnostic-details}) yields a potential explanation: GLMOCR produces 
substantially more formatting overhead, which region-based matching against the JSON 
annotations does not heavily penalise, whereas RolmOCR's segment boundaries potentially diverge from 
the annotated blocks, which can cost it under JSON matching but not when the page is compared as 
Markdown. Neither ranking has been shown to be wrong; each reflects a different, equally defensible 
assumption about what constitutes correct output.

For practitioners, a profile of scores across reasonable design choices is therefore more informative than a 
single ordering, particularly when the differences can be read against candidate failure modes: it 
indicates which configurations a model performs well under, and so which document types and 
failure tolerances it suits.

\subsection{Segment-Level Matching Amplifies the Effect of the Ground-Truth Format}

The prediction segments are identical under both ground-truth formats; what differs is how the 
reference is divided. The Markdown rendering merges, omits and rewrites some annotated blocks, whereas the JSON format excludes eleven categories from the text score. Segment-to-segment matching has 
no slack for such differences, since a correctly parsed segment paired with the wrong reference 
can score near zero, whereas concatenating the page removes the pairing step entirely. This is the 
leaderboard-level consequence of the matching-strategy bias identified by the OmniDocBench 
authors~\cite{mineru25}.

MGAM targets precisely this step, searching for an optimal segmentation granularity on the 
prediction side while holding the ground truth fixed. Even with MGAM applied, however, the two 
ground-truth formats agree at only $\tau = 0.400$ under \texttt{quick\_match}, so the choice of 
reference format remains consequential after the redesign.

\subsection{Limitations and Future Work}

\paragraph{Validation of the diagnostic protocol.}
The mapping from signature families to failure modes in
Section~\ref{sec:diagnostic} is derived from score-signature clustering,
aggregate text statistics, and targeted manual review of exemplar pages. It is
not validated against a held-out labelled sample, so we report the mechanisms
behind each family but not the precision with which a signature predicts the
dominant failure mode on an unseen page. An annotation study, labelling a
sample of pages with their dominant failure mode and measuring agreement with
the predicted signature family, would turn the protocol from a descriptive
mapping into an instrument with known error rates. Relatedly, our
characterisation covers the most frequent families; the long tail of the
signature vocabulary (301 signatures, of which the top 20 account for 78\% of
rows) remains unexamined.

\paragraph{Benchmark coverage.}
Our study is centred on OmniDocBench v1.5, which to our knowledge is the only
document parsing benchmark exposing configurable alignment; benchmarks outside
its ecosystem fix a single alignment procedure, or bypass it altogether, so the controlled comparison
cannot currently be run elsewhere. Extending RankShift to inheritor benchmarks
that retain both alignment options (e.g.\ MDPBench, MPDocBench-Parse) would
test whether the relative importance of alignment, dataset, and metric effects
holds under different corpus compositions. This bears directly on our
diagnostic protocol: the E2E-favouring signatures that dominate here reflect
the document types present in OmniDocBench, and a math- or slide-heavy corpus
should shift the balance toward the M2M-favouring families we observe only
rarely. That the comparison is unavailable elsewhere is itself an argument for
exposing alignment as an explicit, configurable variable.

\section{Conclusion}

Document parsing benchmarks are used to compare model performance, but rankings are conditional on evaluation design. RankShift shows that these design choices do not contribute equally: alignment 
strategy and dataset composition substantially reorder model rankings, while metric choice leaves 
them largely intact. Disagreement across alignment strategies is also potentially informative, since a model's profile 
of scores may indicate which failure modes dominate its output. We therefore propose that benchmark releases report the evaluation configuration 
alongside results, together with rank-stability profiles under reasonable design choices.

\FloatBarrier
\clearpage

{\small
\bibliographystyle{plainnat}
\bibliography{bib/references}

@article{niu2025mineru25,
  title        = {{MinerU2.5}: A Decoupled Vision-Language Model for Efficient High-Resolution Document Parsing},
  author       = {Niu, Junbo and others},
  journal      = {arXiv preprint arXiv:2509.22186},
  year         = {2025},
  url          = {https://arxiv.org/abs/2509.22186}
}

@article{mineru25,
  title        = {{MinerU2.5-Pro}: Pushing the Limits of Data-Centric Document Parsing at Scale},
  author       = {Wang, Bin and others},
  journal      = {arXiv preprint arXiv:2604.04771},
  year         = {2026},
  doi          = {10.48550/arXiv.2604.04771}
}

@article{hunyuan2025ocr,
  title        = {{HunyuanOCR Technical Report}},
  author       = {{Hunyuan Vision Team} and others},
  journal      = {arXiv preprint arXiv:2511.19575},
  year         = {2025},
  url          = {https://arxiv.org/abs/2511.19575}
}

@article{cao2026youtuparsing,
  title        = {{Youtu-Parsing: Perception, Structuring and Recognition via High-Parallelism Decoding}},
  author       = {Cao, Haoyu and others},
  journal      = {arXiv preprint arXiv:2601.20430},
  year         = {2026},
  url          = {https://arxiv.org/abs/2601.20430}
}

@article{li2025dotsocr,
  title        = {{dots.ocr: Multilingual Document Layout Parsing in a Single Vision-Language Model}},
  author       = {Li, Yumeng and Yang, Guang and Liu, Hao and Wang, Bowen and Zhang, Colin},
  journal      = {arXiv preprint arXiv:2512.02498},
  year         = {2025},
  url          = {https://arxiv.org/abs/2512.02498}
}

@article{olmocr,
  title        = {olmOCR: Unlocking Trillions of Tokens in PDFs with Vision Language Models},
  author       = {Poznanski, Jake and Borchardt, Jon and Dunkelberger, Jason and Huff, Regan and Lin, Daniel and Rangapur, Aman and Wilhelm, Christopher and Lo, Kyle and Soldaini, Luca},
  journal      = {arXiv preprint arXiv:2502.18443},
  year         = {2025}
}

@article{parsebench,
  title        = {{ParseBench}: A Document Parsing Benchmark for AI Agents},
  author       = {Zhang, Boyang and Acosta, Sebasti{\'a}n G. and Carlson, Preston and Bron, Sacha and Doulcet, Pierre-Lo{\"i}c and Ospina, Daniel B. and Suo, Simon},
  journal      = {arXiv preprint arXiv:2604.08538},
  year         = {2026},
  doi          = {10.48550/arXiv.2604.08538}
}

@inproceedings{ouyang2024omnidocbench,
  title        = {{OmniDocBench}: Benchmarking Diverse PDF Document Parsing with Comprehensive Annotations},
  author       = {Ouyang, Linke and Qu, Yuan and Zhou, Hongbin and Zhu, Jiawei and Zhang, Rui and Lin, Qunshu and Wang, Bin and Zhao, Zhiyuan and Jiang, Man and Zhao, Xiaomeng and Shi, Jin and Wu, Fan and Chu, Pei and Liu, Minghao and Li, Zhenxiang and Xu, Chao and Zhang, Bo and Shi, Botian and Tu, Zhongying and He, Conghui},
  booktitle    = {Proceedings of the IEEE/CVF Conference on Computer Vision and Pattern Recognition},
  year         = {2025}
}

@article{zhang2025docrinspector,
  title        = {{DOCR-Inspector}: Fine-Grained and Automated Evaluation of Document Parsing with {VLM}},
  author       = {Zhang, Qintong and others},
  journal      = {arXiv preprint arXiv:2512.10619},
  year         = {2025}
}

@article{mdpbench2026,
  title        = {{MDPBench}: A Benchmark for Multilingual Document Parsing in Real-World Scenarios},
  author       = {Li, Zhang and Lin, Zhibo and Liu, Qiang and Zhang, Ziyang and Zhang, Shuo and Guo, Zidun and Song, Jiajun and Zhang, Jiarui and Bai, Xiang and Liu, Yuliang},
  journal      = {arXiv preprint arXiv:2603.28130},
  year         = {2026}
}

@article{mpdocbenchparse2026,
  title        = {{MPDocBench-Parse}: Benchmarking Practical Multi-page Document Parsing},
  author       = {Zhou, Bangbang and Xing, Hangdi and Chen, Yifan and Xu, Jianjun and Zheng, Qi and Gao, Feiyu and Yang, Zhibo and Bai, Shuai and Yan, Ming and Ye, Jieping and Xie, Hongtao},
  journal      = {arXiv preprint arXiv:2605.22100},
  year         = {2026}
}

@inproceedings{induocrbench2026,
  title        = {When Good {OCR} Is Not Enough: Benchmarking {OCR} Robustness for Retrieval-Augmented Generation},
  author       = {Sun, Lin and Wang, Dexian and Huang, Jingang and Zhang, Linglin and Jia, Change and Cheng, Zhengwei and Zhang, Xiangzheng},
  booktitle    = {Proceedings of the 64th Annual Meeting of the Association for Computational Linguistics (Volume 6: Industry Track)},
  pages        = {884--894},
  publisher    = {Association for Computational Linguistics},
  doi          = {10.18653/v1/2026.acl-industry.60},
  year         = {2026}
}

@article{drdocbench2026,
  title        = {Dr. {DocBench}: A Comprehensive Benchmark for Expert-Level and Difficult Document Parsing},
  author       = {Yang, Minglai and Yu, {Xinyan Velocity} and Li, Pengyuan and Guo, Xinyu and Qi, Zhenting and Kim, Konwoo and Ye, Longtian and Luo, Xiaolong and Bi, Jinhe and Zhang, Henry and Riaz, Haris and Zhang, Xuan and Xiao, Yunze and Liu, Bangya and Tang, Tom and Zhao, Yunfei and Lin, Qunshu and Wang, Zihan and Liu, Minghao and Li, {Michael Lingzhi} and Du, Yilun and Thomason, Jesse and Feris, Rogerio and Pentland, Alex and He, Zexue},
  journal      = {arXiv preprint arXiv:2606.01393},
  year         = {2026}
}

@inproceedings{wang2025imageovertext,
  title        = {Image Over Text: Transforming Formula Recognition Evaluation with Character Detection Matching},
  author       = {Wang, Bin and Wu, Fan and Ouyang, Linke and Gu, Zhuangcheng and Zhang, Rui and Xia, Renqiu and Shi, Botian and Zhang, Bo and He, Conghui},
  booktitle    = {Proceedings of the IEEE/CVF Conference on Computer Vision and Pattern Recognition (CVPR)},
  month        = jun,
  year         = {2025},
  pages        = {19681--19690},
  doi          = {10.1109/CVPR52734.2025.01833},
  url          = {https://ieeexplore.ieee.org/document/11094480}
}

@inproceedings{post2018call,
  title        = {A Call for Clarity in Reporting {BLEU} Scores},
  author       = {Post, Matt},
  booktitle    = {Proceedings of the Third Conference on Machine Translation: Research Papers},
  year         = {2018}
}

@article{marie2021scientific,
  title        = {Scientific Credibility of Machine Translation Research: A Meta-Evaluation of 769 Papers},
  author       = {Marie, Benjamin and Fujita, Atsushi and Rubino, Raphael},
  journal      = {arXiv preprint arXiv:2106.15195},
  year         = {2021},
  doi          = {10.48550/arXiv.2106.15195}
}

@article{dehghani2021benchmarklottery,
  title        = {The Benchmark Lottery},
  author       = {Dehghani, Mostafa and Tay, Yi and Gritsenko, Alexey A. and Zhao, Zhe and Houlsby, Neil and Diaz, Fernando and Metzler, Donald and Vinyals, Oriol},
  journal      = {arXiv preprint arXiv:2107.07002},
  year         = {2021},
  doi          = {10.48550/arXiv.2107.07002}
}

@article{zhou2025real5,
  title        = {{Real5-OmniDocBench}: A Full-Scale Physical Reconstruction Benchmark for Robust Document Parsing in the Wild},
  author       = {Zhou, Changda and others},
  journal      = {arXiv preprint arXiv:2603.04205},
  year         = {2026}
}

@misc{docling2024,
  title        = {Docling: Technical Report},
  author       = {{IBM Research and Contributors}},
  howpublished = {GitHub repository and technical report},
  year         = {2024}
}

@article{wei2024gotocr,
  title        = {{General OCR Theory: Towards OCR-2.0 via a Unified End-to-end Model}},
  author       = {Wei, Haoran and others},
  journal      = {arXiv preprint arXiv:2409.01704},
  year         = {2024}
}

@article{glmocr2026,
  title        = {{GLM-OCR} Technical Report},
  author       = {Duan, Shuaiqi and others},
  journal      = {arXiv preprint arXiv:2603.10910},
  year         = {2026}
}

@article{cui2025paddleocrvl,
  title        = {{PaddleOCR-VL}-1.5: Towards a Multi-Task 0.9B {VLM} for Robust In-the-Wild Document Parsing},
  author       = {Cui, Cheng and others},
  journal      = {arXiv preprint arXiv:2601.21957},
  year         = {2026}
}

@article{feng2025dolphin,
  title        = {Dolphin: Document Image Parsing via Heterogeneous Anchor Prompting},
  author       = {Feng, Hao and Wei, Shu and Fei, Xiang and Shi, Wei and Han, Yingdong and Liao, Lei and Lu, Jinghui and Wu, Binghong and Liu, Qi and Lin, Chunhui and Tang, Jingqun and Liu, Hao and Huang, Can},
  journal      = {arXiv preprint arXiv:2505.14059},
  year         = {2025}
}

@article{li2025monkeyocr,
  title        = {MonkeyOCR: Document Parsing with a Structure-Recognition-Relation Triplet Paradigm},
  author       = {Li, Zhang and others},
  journal      = {arXiv preprint arXiv:2506.05218},
  year         = {2025}
}

@article{deepseekocr2025,
  title        = {{DeepSeek-OCR} 2: Visual Causal Flow},
  author       = {Wei, Haoran and others},
  journal      = {arXiv preprint arXiv:2601.20552},
  year         = {2026}
}

@misc{chandra2025,
  title        = {Chandra OCR 2},
  author       = {{Datalab}},
  howpublished = {Hugging Face model card},
  year         = {2025}
}

@misc{rolmocr2025,
  author       = {Reducto AI},
  title        = {RolmOCR: A Faster, Lighter Open Source OCR Model},
  howpublished = {Hugging Face model card},
  year         = {2025}
}

@article{kwon2023vllm,
  title        = {Efficient Memory Management for Large Language Model Serving with {PagedAttention}},
  author       = {Kwon, Woosuk and Li, Zhuohan and Zhuang, Siyuan and Sheng, Ying and Zheng, Lianmin and Yu, Cody Hao and Gonzalez, Joseph E. and Zhang, Hao and Stoica, Ion},
  journal      = {arXiv preprint arXiv:2309.06180},
  year         = {2023},
  doi          = {10.48550/arXiv.2309.06180}
}

@misc{openai2025,
  title        = {{GPT}-5.5 model documentation},
  author       = {{OpenAI}},
  howpublished = {OpenAI developer documentation},
  year         = {2026}
}

@article{comanici2025gemini,
  title        = {Gemini 2.5: Pushing the Frontier with Advanced Reasoning, Multimodality, Long Context, and Next Generation Agentic Capabilities},
  author       = {Comanici, Gheorghe and Bieber, Eric and Schaekermann, Mike and Pasupat, Ice and Sachdeva, Noveen and Dhillon, Inderjit and Blistein, Marcel and Ram, Ori and Zhang, Dan and Rosen, Evan and others},
  journal      = {arXiv preprint arXiv:2507.06261},
  year         = {2025},
  doi          = {10.48550/arXiv.2507.06261}
}

@misc{google2026gemini31flash,
  title        = {Gemini 3.1 Flash Live Preview},
  author       = {{Google}},
  howpublished = {Google AI for Developers model documentation},
  year         = {2026},
  url          = {https://ai.google.dev/gemini-api/docs/models/gemini-3.1-flash-live-preview},
  note         = {Accessed 2026-08-29}
}

@inproceedings{mathur2020tangled,
  title        = {Tangled up in {BLEU}: Reevaluating the Evaluation of Automatic Machine Translation Evaluation Metrics},
  author       = {Mathur, Nitika and Baldwin, Timothy and Cohn, Trevor},
  booktitle    = {Proceedings of the 58th Annual Meeting of the Association for Computational Linguistics},
  pages        = {4984--4997},
  year         = {2020}
}

@inproceedings{taori2020measuring,
  title        = {Measuring Robustness to Natural Distribution Shifts in Image Classification},
  author       = {Taori, Rohan and Dave, Achal and Shankar, Vaishaal and others},
  booktitle    = {Advances in Neural Information Processing Systems},
  year         = {2020}
}

@article{alzahrani-etal-2024-benchmarks,
  title        = {When Benchmarks are Targets: Revealing the Sensitivity of Large Language Model Leaderboards},
  author       = {Alzahrani, Norah A. and Alyahya, Hisham Abdullah and Alnumay, Yazeed and Alrashed, Sultan and Alsubaie, Shaykhah Z. and Almushayqih, Yousef and Mirza, Faisal Abdulrahman and Alotaibi, Nouf M. and Al-Twairesh, Nora and Alowisheq, Areeb and Bari, M. Saiful and Khan, Haidar},
  journal      = {arXiv preprint arXiv:2402.01781},
  year         = {2024},
  doi          = {10.48550/arXiv.2402.01781}
}

@article{cais2026hle,
  title        = {A benchmark of expert-level academic questions to assess {AI} capabilities},
  author       = {{Center for AI Safety} and {Scale AI} and {HLE Contributors Consortium}},
  journal      = {Nature},
  volume       = {649},
  pages        = {1139--1146},
  year         = {2026},
  doi          = {10.1038/s41586-025-09962-4}
}

@article{qian2026benchmark2,
  title        = {{Benchmark$^2$}: Systematic Evaluation of LLM Benchmarks},
  author       = {Qian, Qi and Huang, Chengsong and Xu, Jingwen and Lv, Changze and Wu, Muling and Liu, Wenhao and Wang, Xiaohua and Wang, Zhenghua and Huang, Zisu and Tian, Muzhao and Xu, Jianhan and Hu, Kun and Wang, He-Da and Hu, Yao and Huang, Xuanjing and Zheng, Xiaoqing},
  journal      = {arXiv preprint arXiv:2601.03986},
  year         = {2026}
}

@article{kendiukhov2026grn,
  title        = {Quantifying Ranking Instability Across Evaluation Protocol Axes in Gene Regulatory Network Benchmarking},
  author       = {Kendiukhov, Ihor},
  journal      = {arXiv preprint arXiv:2603.03493},
  year         = {2026},
  doi          = {10.48550/arXiv.2603.03493}
}

@article{li2026safetyrepro,
  title        = {{SafetyRepro: Configuration-Conditional Rank Instability on Alignment Benchmarks}},
  author       = {Li, Yanhang and Fan, Zhichao and Zhuang, Zexin},
  journal      = {arXiv preprint arXiv:2605.25492},
  year         = {2026},
  doi          = {10.48550/arXiv.2605.25492}
}

@article{eriksson2025trust,
  title        = {Can We Trust {AI} Benchmarks? An Interdisciplinary Review of Current Issues in {AI} Evaluation},
  author       = {Eriksson, Maria and Purificato, Erasmo and Noroozian, Arman and Vinagre, Jo{\~a}o and Chaslot, Guillaume and G{'o}mez, Emilia and Fernandez-Llorca, David},
  journal      = {arXiv preprint arXiv:2502.06559},
  year         = {2025}
}
}

\clearpage
\FloatBarrier
\appendix
\renewcommand{\thesection}{Appendix \Alph{section}:}
\renewcommand{\thesubsection}{\thesection\arabic{subsection}}

\section{Additional Tables}

Supporting tables for the results reported in the main text are provided below.

\renewcommand{\thetable}{A\arabic{table}}
\setcounter{table}{0}

\begin{table}[H]
\centering
\caption{Dataset configurations used in Experiment 1. Partition-based comparisons use disjoint page subsets, while the capture-condition comparison pairs two versions of the same pages.}
\label{tab:dataset-configs}
\small
\begin{tabular}{@{}>{\raggedright\arraybackslash}p{0.13\textwidth}>{\raggedright\arraybackslash}p{0.43\textwidth}>{\raggedright\arraybackslash}p{0.17\textwidth}>{\raggedright\arraybackslash}p{0.17\textwidth}@{}}
\toprule
Variation & Configurations & Comparisons & Evidence relationship \\
\midrule
Page type
& PPT2PDF; academic literature; book; colorful textbook; exam paper; magazine; newspaper; note; research report
& All pairwise ($36$)
& Disjoint page subsets \\
Layout
& 1$+$ column; double column; other layout; single column; three column
& All pairwise ($10$)
& Disjoint page subsets \\
Capture condition
& OmniDocBench v1.5 digital pages; Real5 Scanning reconstructions
& Digital vs.\ scanned ($1$)
& Paired versions of the same 1,355 pages \\
\bottomrule
\end{tabular}
\end{table}

\begin{table}[H]
\centering
\caption{Experiment 1 model rankings on OmniDocBench v1.5 vs.\ Real5 Scanning ($n=16$ models, E2E \texttt{quick\_match} text NED). $\Delta = {}$OmniDocBench rank $-$ Real5 rank (1 = best); negative $\Delta$ means a worse rank under Real5.}
\label{tab:experiment1-real5-rank-shift-text-ned}
\footnotesize
\begin{tabular}{lrrr}
\toprule
Model & OmniDocBench rank & Real5 rank & $\Delta$ Real5 position \\
\midrule
dots.ocr & \textbf{1} & 6 & $-5$ \\
GLMOCR & 2 & 2 & 0 \\
PaddleOCR-VL-1.5 & 3 & \textbf{1} & $+2$ \\
Youtu-Parsing & 4 & 4 & 0 \\
MinerU2.5 & 5 & 3 & $+2$ \\
DeepSeek-OCR 2 & 6 & 5 & $+1$ \\
Chandra 2 & 7 & 7 & 0 \\
Hunyuan OCR & 8 & 9 & $-1$ \\
Gemini 3.1 Flash & 9 & 8 & $+1$ \\
Gemini 2.5 Pro & 10 & 11 & $-1$ \\
Dolphin 1.5 & 11 & 10 & $+1$ \\
MonkeyOCR-pro-3B & 12 & 14 & $-2$ \\
GPT-5.5 & 13 & 13 & 0 \\
RolmOCR & 14 & 12 & $+2$ \\
Docling & 15 & 15 & 0 \\
GOT-OCR2.0 & 16 & 16 & 0 \\
\bottomrule
\end{tabular}
\end{table}

\clearpage
\subsection{Statistical Significance of Rank Reversals}

For Experiment 2 we report formal significance testing at three complementary levels, each with Benjamini--Hochberg FDR correction at $\alpha=0.05$.

\begin{table}[H]
\centering
\footnotesize
\setlength{\tabcolsep}{4pt}
\caption{Alignment-pair rank reversal rates ($n=16$ models, 120 model pairs per alignment pair). One-sided binomial tests vs.\ within-E2E baseline ($\sim16.7\%$); $q$-values from Benjamini--Hochberg FDR over 15 alignment pairs ($\alpha=0.05$). 95\% CIs from page bootstrap. An asterisk indicates significance after FDR correction ($q<0.05$).}
\label{tab:reversal_significance}
\begin{tabular}{p{6.8cm} r p{2.4cm} r}
\toprule
\textbf{Comparison} & \textbf{Rate} & \textbf{95\% CI} & \textbf{$q$-value} \\
\midrule
\multicolumn{4}{l}{\textit{Cross-paradigm (E2E $\leftrightarrow$ M2M)}} \\
E2E quick\_match vs.\ M2M simple\_match & 35.8\% & [30.0\%, 39.2\%] & $4.8\times10^{-6}$\textsuperscript{*} \\
E2E simple\_match vs.\ M2M simple\_match & 35.0\% & [27.5\%, 36.7\%] & $6.8\times10^{-6}$\textsuperscript{*} \\
E2E simple\_match vs.\ M2M quick\_match & 30.0\% & [23.3\%, 32.5\%] & 0.001\textsuperscript{*} \\
E2E simple\_match vs.\ M2M no\_split & 29.2\% & [21.7\%, 31.7\%] & 0.002\textsuperscript{*} \\
E2E quick\_match vs.\ M2M no\_split & 28.3\% & [21.7\%, 31.7\%] & 0.003\textsuperscript{*} \\
E2E quick\_match vs.\ M2M quick\_match & 25.8\% & [20.8\%, 29.2\%] & 0.018\textsuperscript{*} \\
E2E no\_split vs.\ M2M simple\_match & 20.0\% & [15.8\%, 26.7\%] & 0.363 \\
E2E no\_split vs.\ M2M quick\_match & 12.5\% & [8.3\%, 17.5\%] & 0.986 \\
E2E no\_split vs.\ M2M no\_split & 10.0\% & [5.8\%, 15.8\%] & 0.986 \\
\midrule
\multicolumn{4}{l}{\textit{Within E2E}} \\
E2E quick\_match vs.\ E2E simple\_match & 11.7\% & [7.5\%, 14.2\%] & 0.986 \\
E2E quick\_match vs.\ E2E no\_split & 20.0\% & [15.0\%, 25.0\%] & 0.363 \\
E2E simple\_match vs.\ E2E no\_split & 18.3\% & [14.2\%, 23.3\%] & 0.475 \\
\midrule
\multicolumn{4}{l}{\textit{Within M2M}} \\
M2M quick\_match vs.\ M2M simple\_match & 16.7\% & [15.8\%, 24.2\%] & 0.372 \\
M2M quick\_match vs.\ M2M no\_split & 9.2\% & [6.7\%, 14.2\%] & 0.986 \\
M2M simple\_match vs.\ M2M no\_split & 15.8\% & [13.3\%, 20.8\%] & 0.459 \\
\bottomrule
\end{tabular}
\end{table}

\paragraph{Alignment-pair reversal rate.}
For each pair of alignments, we count how often two models' relative order on a page reverses between the alignments (120 model pairs per alignment pair for 16 models) and test whether cross-paradigm reversal rates exceed the mean within-E2E baseline ($\sim16.7\%$), which captures normal ranking noise under a fixed protocol. We use one-sided binomial tests with BH-FDR correction over the 15 alignment pairs. Six of the nine cross-paradigm (E2E$\leftrightarrow$M2M) pairs show significantly elevated reversal rates (25--36\% vs.\ the $\sim17\%$ within-protocol baseline); none of the six within-paradigm pairs do. Rank instability is therefore protocol-specific rather than general noise. All three nonsignificant cross-paradigm comparisons involve E2E \texttt{no\_split}; the lowest rate, E2E \texttt{no\_split} vs.\ M2M \texttt{no\_split} (10\%), is expected because both alignments use whole-page matching, so the paradigm distinction largely collapses.

\paragraph{Per-model rank displacement.}
Page-bootstrap tests ($2{,}000$ resamples) confirm that individual models shift rank systematically across alignments: 78 model$\times$alignment-pair displacements are FDR-significant, including 51 under cross E2E$\leftrightarrow$M2M pairs. For example, RolmOCR improves by eleven ranks under M2M \texttt{no\_split} vs.\ E2E \texttt{quick\_match}, whereas GLMOCR worsens by eight ranks under the same pair.

\paragraph{Model-pair page flips.}
At page granularity, 869 of 1,800 model-pair$\times$alignment-pair combinations show significant head-to-head flip rates under two-sided binomial tests against the pair-mean baseline.

Together, these tests establish that the reversals are systematic and statistically robust rather than artifacts of indistinguishable neighbouring models.

\clearpage
\section{Additional Figures}

This appendix presents supplementary figures for the three instability experiments.

\renewcommand{\thefigure}{B\arabic{figure}}
\setcounter{figure}{0}

\begin{figure}[H]
\centering
\includegraphics[width=\textwidth]{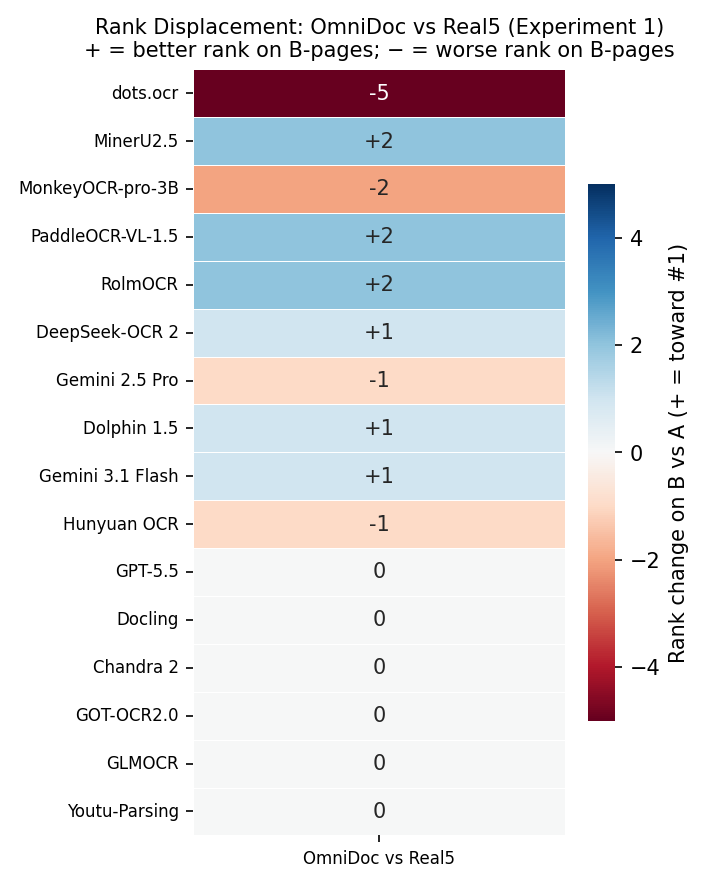}
\caption{Cross-dataset rank displacement from OmniDocBench v1.5 to Real5 Scanning in Experiment~1. Values report OmniDocBench rank minus Real5 rank: positive (blue) values indicate movement toward rank~1 on Real5, negative (red) values indicate a lower rank on Real5, and zero indicates no change.}
\label{fig:appendix-cross-dataset-rank-displacement}
\end{figure}

\begin{figure*}[t]
\centering
\includegraphics[width=\textwidth]{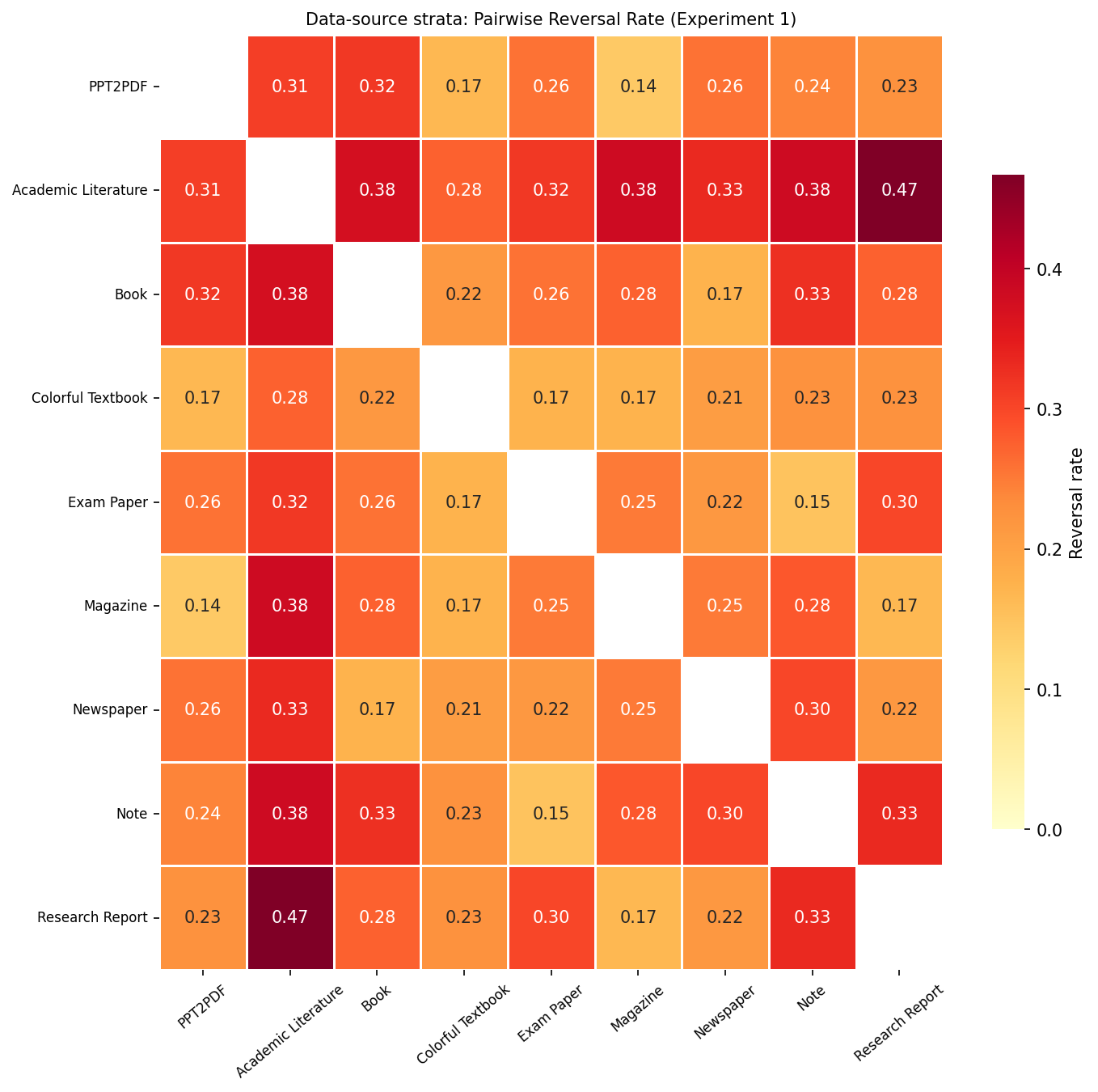}
\caption{Pairwise reversal-rate heatmap for Experiment 1. Each cell gives the share of the 120 
model pairs whose ordering reverses between two of the nine OmniDocBench page-type strata (36 
comparisons).}
\label{fig:appendix-reversal-heatmap-exp1}
\end{figure*}

\begin{figure*}[t]
\centering
\includegraphics[width=\textwidth]{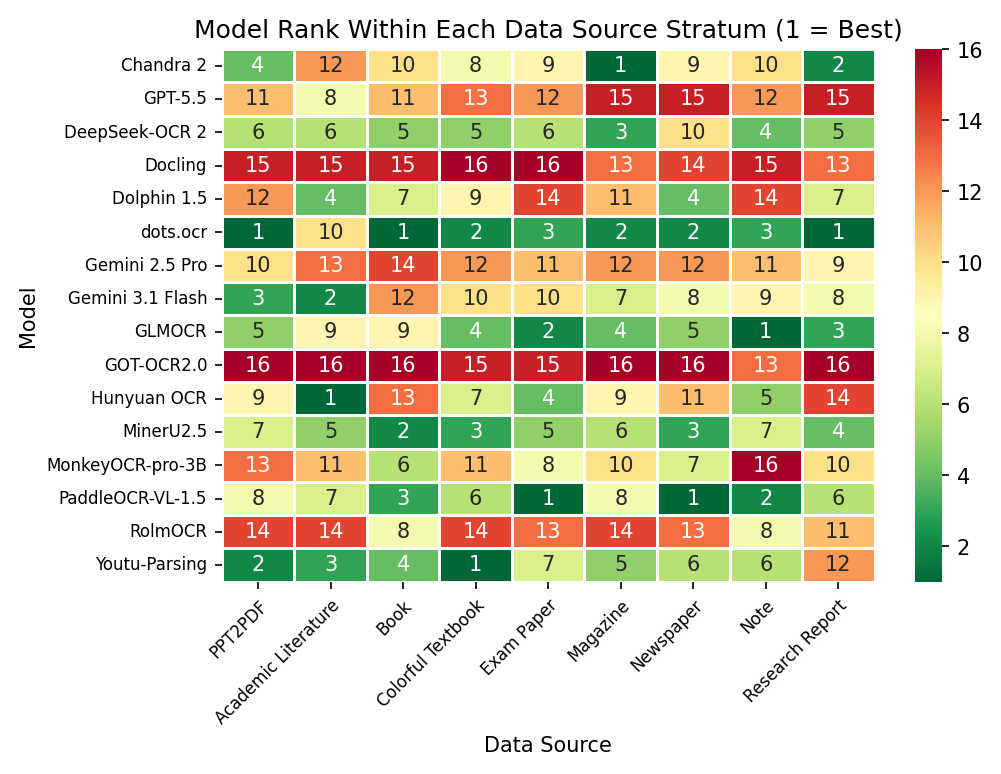}
\caption{Full rank heatmap for Experiment 1: each cell represents a model's rank within each of the nine 
OmniDocBench page-type strata, where 1 is the best-scoring model on that page type and 16 the 
worst.}
\label{fig:appendix-rank-heatmap-data-source}
\end{figure*}

\begin{figure*}[t]
\centering
\includegraphics[width=\textwidth]{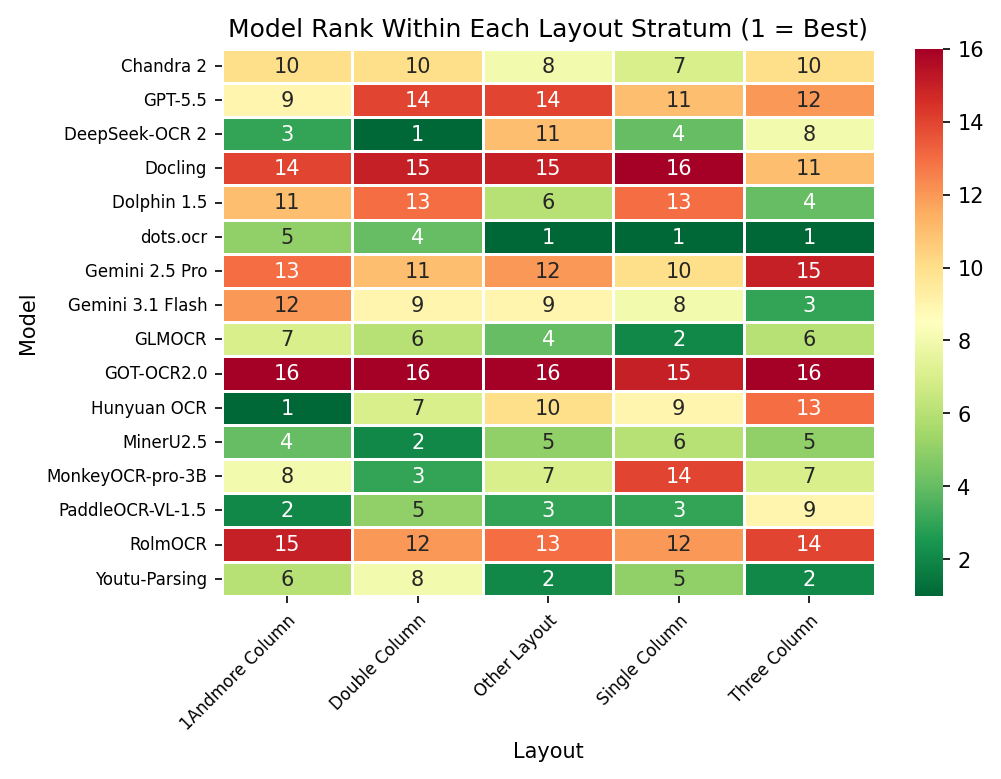}
\caption{Full rank heatmap for Experiment 1: each cell gives a model's rank within one of the five 
OmniDocBench layout strata, where 1 is the best-scoring model on that layout type and 16 the 
worst.}
\label{fig:appendix-rank-heatmap-layout}
\end{figure*}

\begin{figure*}[t]
\centering
\includegraphics[width=\textwidth]{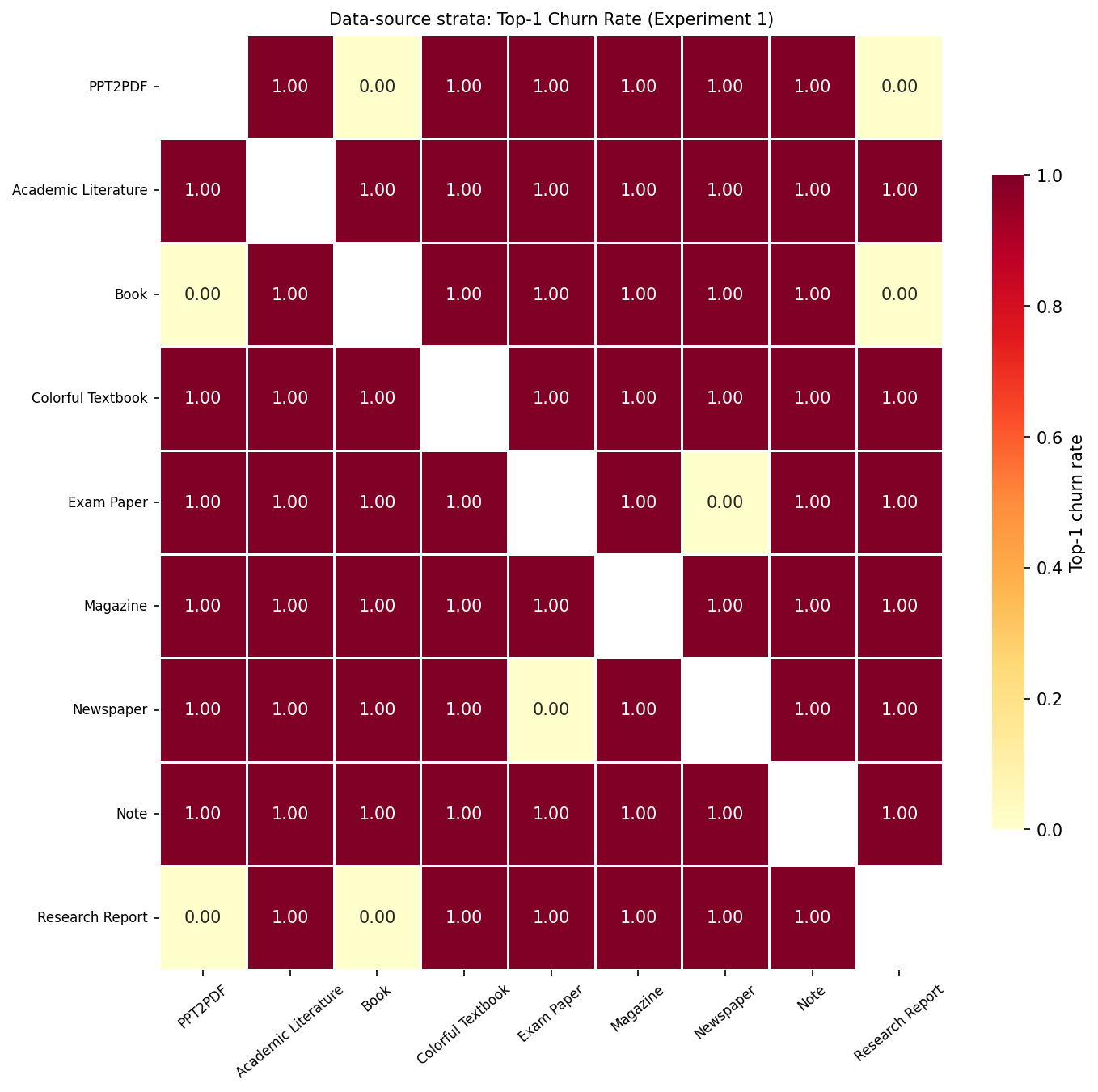}
\caption{Top-1 churn across Experiment 1 comparison settings. Each cell indicates whether the 
highest-scoring model differs between two page-type strata: 1.00 for a different top-ranked model, 
0.00 for the same.}
\label{fig:appendix-topk-churn-k1-exp1}
\end{figure*}

\begin{figure*}[t]
\centering
\includegraphics[width=\textwidth]{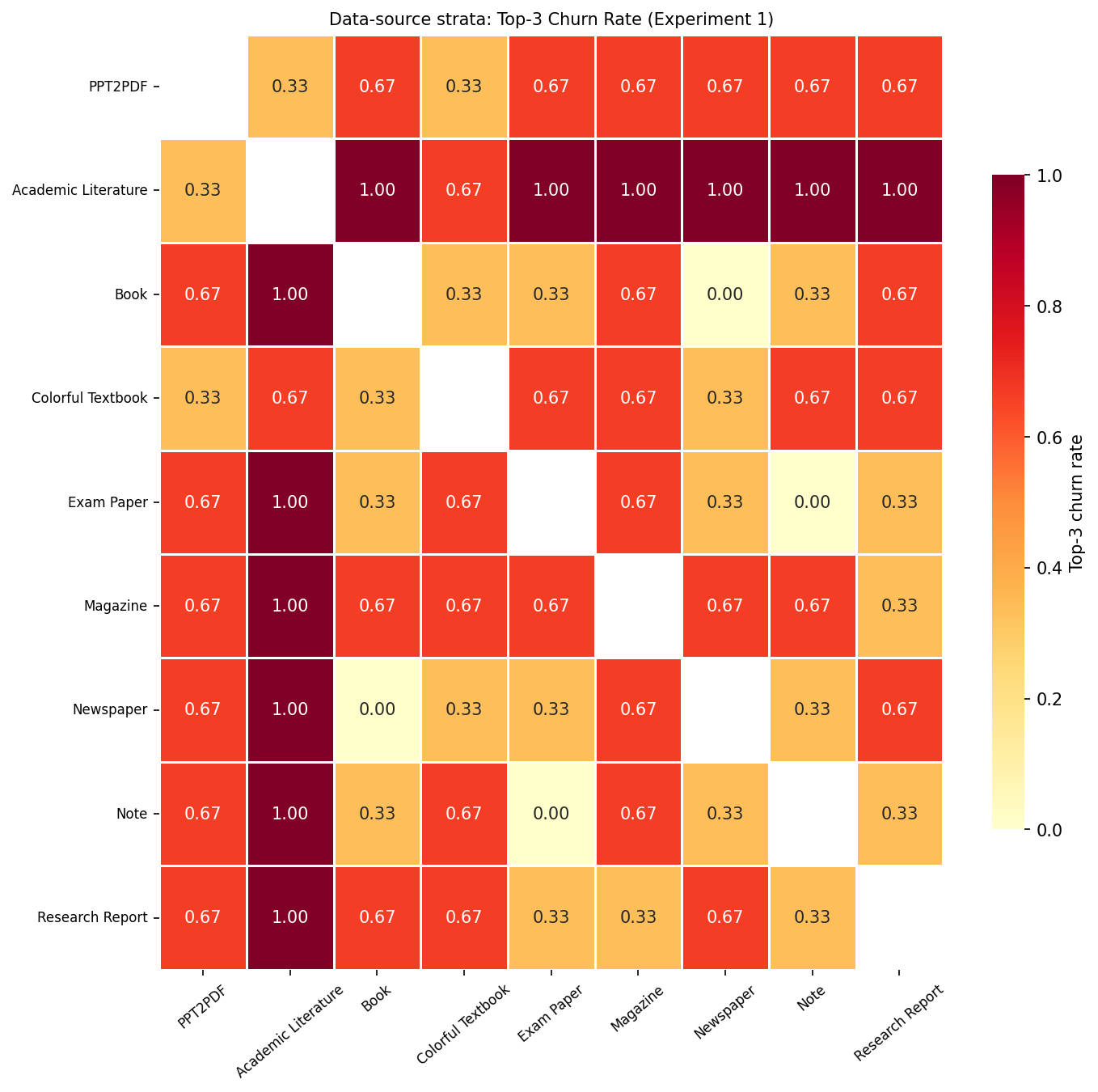}
\caption{Top-3 churn across Experiment 1 comparison settings. Each cell gives the proportion of 
the top-three set that changes between two page-type strata, from 1.00 where none of the three 
models are retained to 0.00 where all top-three models are the same.}
\label{fig:appendix-topk-churn-k3-exp1}
\end{figure*}

\begin{figure*}[t]
\centering
\includegraphics[width=\textwidth]{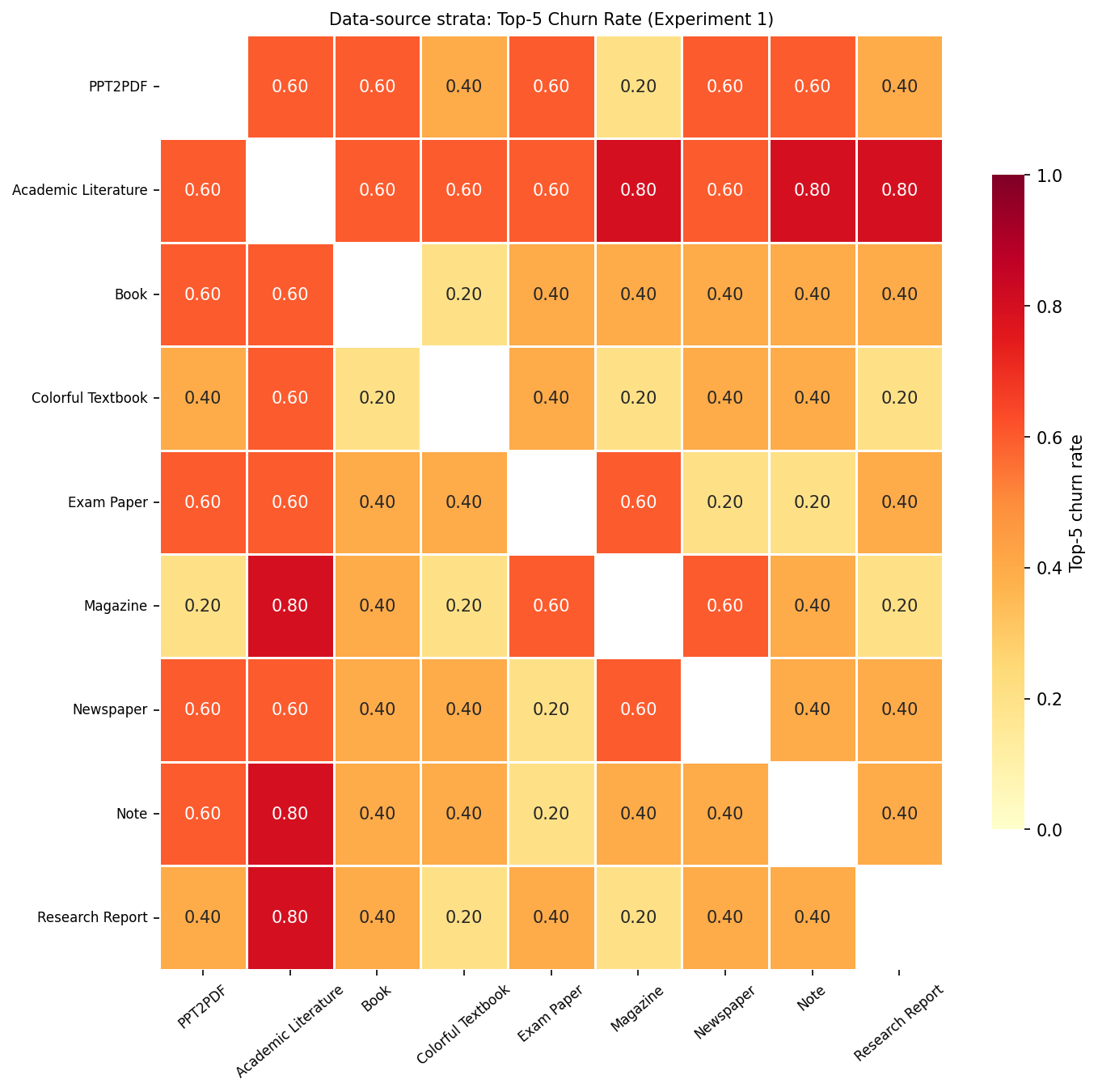}
\caption{Top-5 churn across Experiment~1 comparison settings. Each cell gives the proportion of 
the top-five set that changes between two page-type strata, ranging from 0.20, where one of the 
five models differs, to 0.80, where four differ.}
\label{fig:appendix-topk-churn-k5-exp1}
\end{figure*}

\begin{figure*}[t]
\centering
\includegraphics[width=\textwidth]{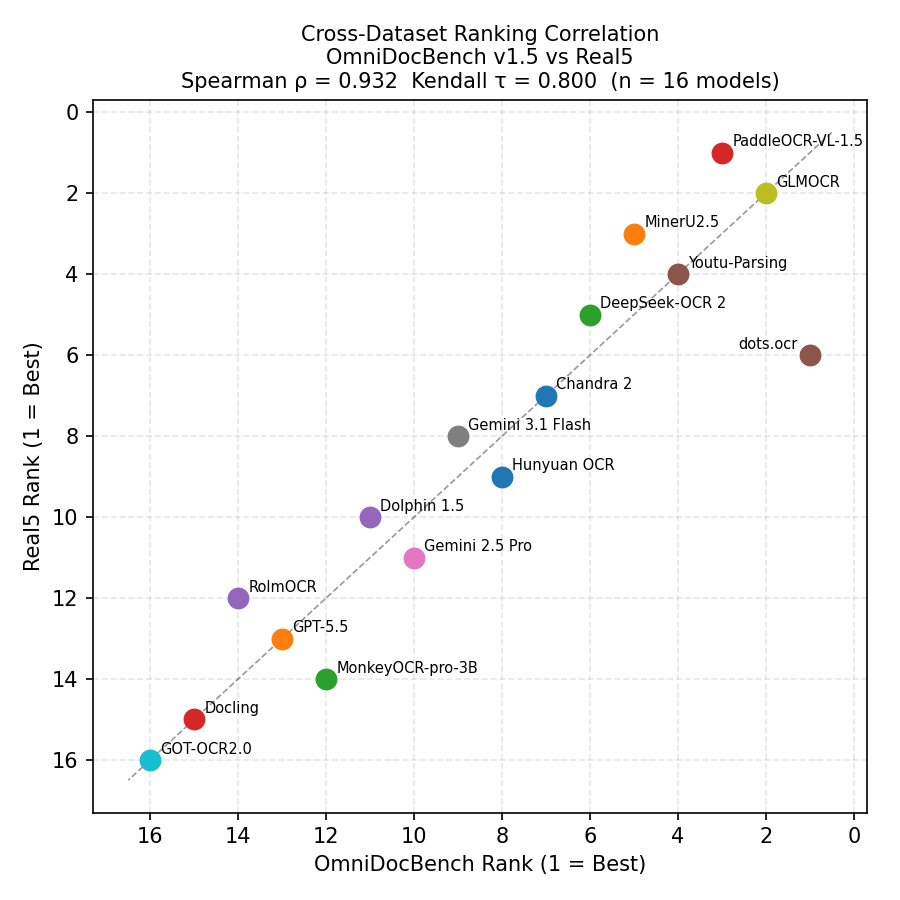}
\caption{Cross-dataset rank scatter for Experiment~1, comparing model ranks on OmniDocBench v1.5 
with the Real5 Scanning subset.}
\label{fig:appendix-cross-dataset-scatter}
\end{figure*}

\begin{figure*}[t]
\centering
\includegraphics[width=\textwidth]{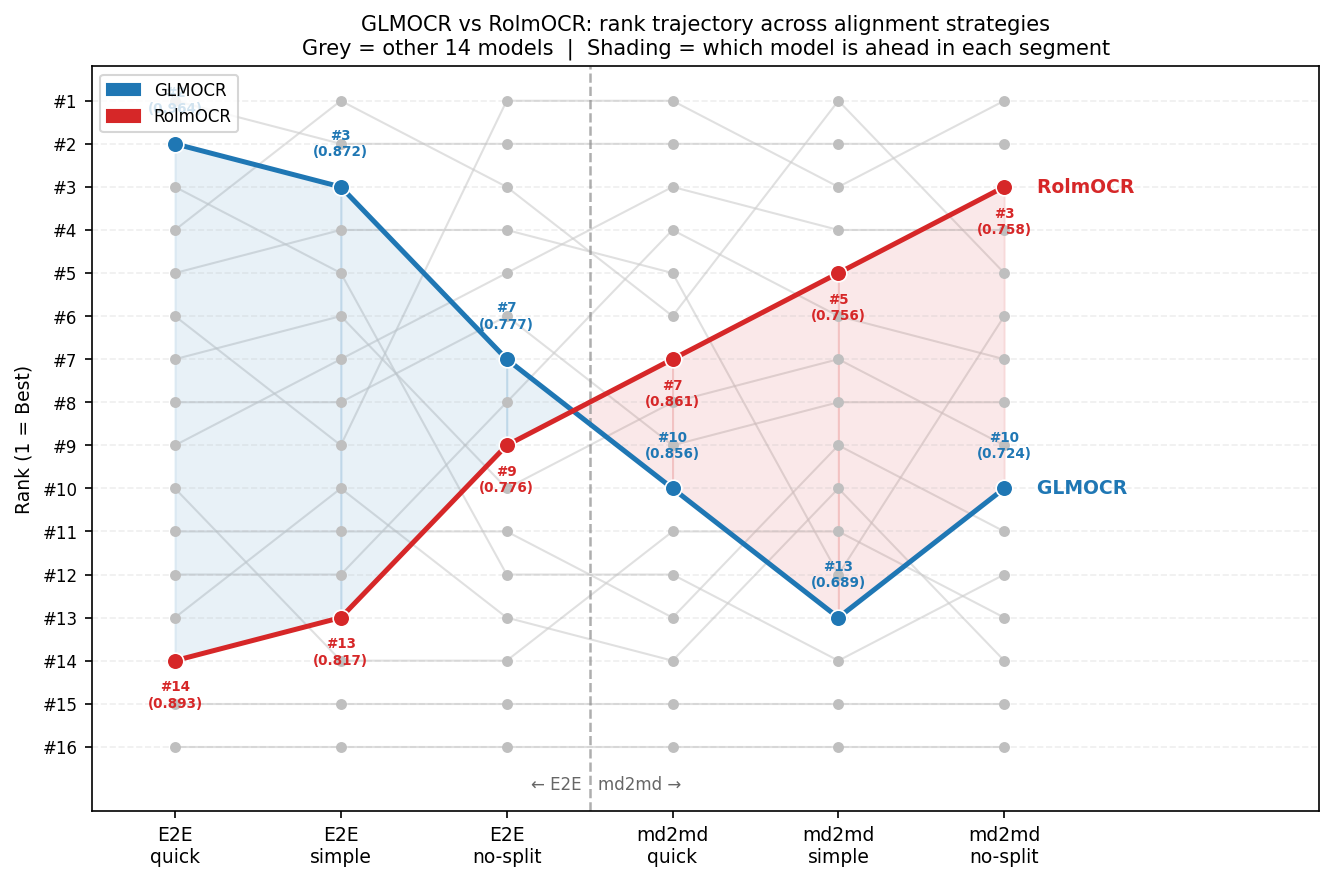}
\caption{Focused bump chart comparing RolmOCR and GLMOCR across the six alignment settings in Experiment~2. GLMOCR ranks higher under every E2E condition and RolmOCR under every M2M condition, 
reversing their relative order.}
\label{fig:appendix-bump-rolmocr-glmocr}
\end{figure*}

\begin{figure*}[t]
\centering
\includegraphics[height=0.78\textheight,keepaspectratio]{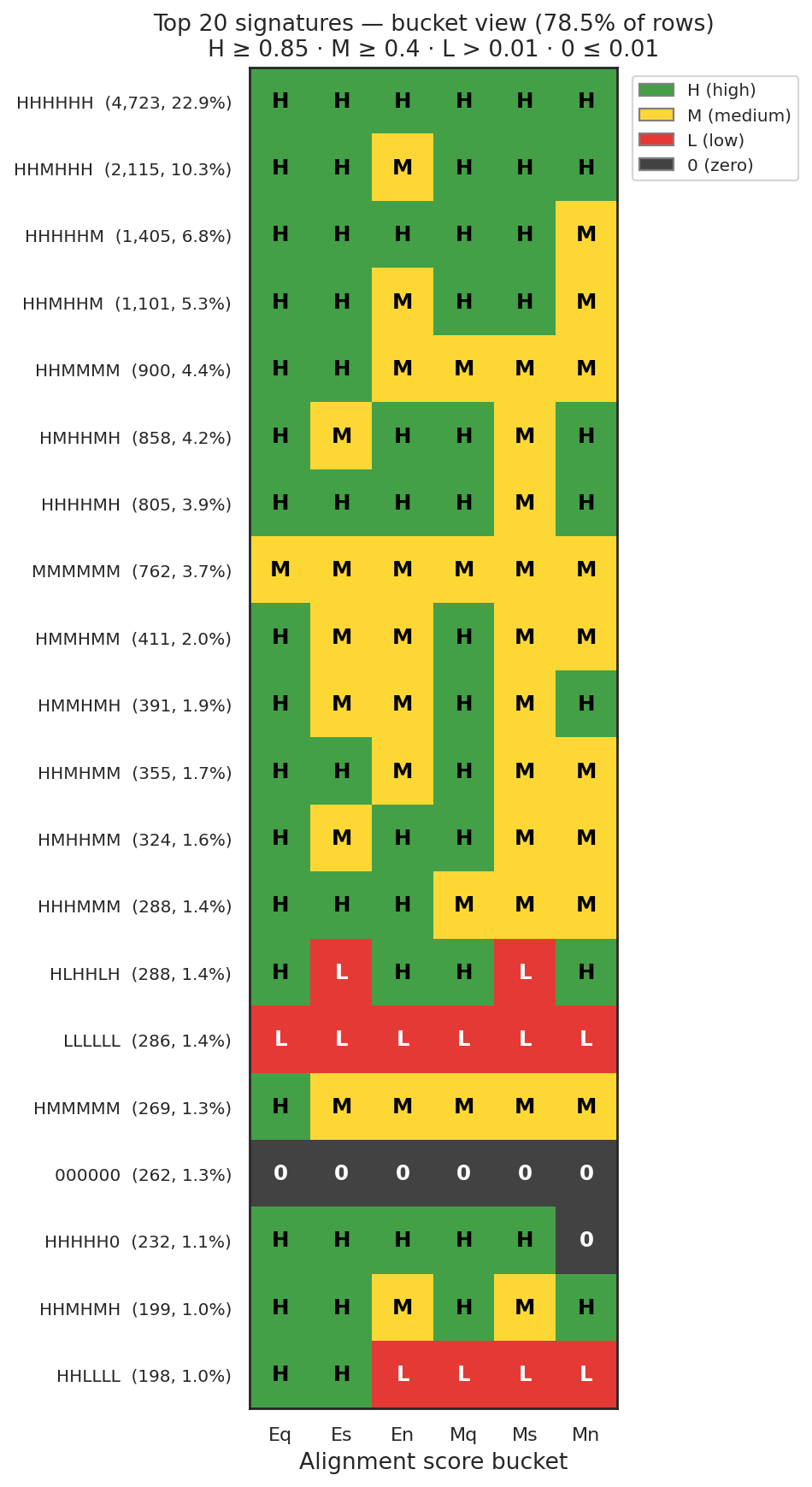}
\caption{Alignment signatures used in the Experiment~2 diagnostic procedure: the 20 most frequent 
patterns, covering 78.5\% of complete page$\times$model rows. Each row is one signature, with 
columns in the fixed order E2E \texttt{quick\_match} ($E_q$), \texttt{simple\_match} ($E_s$) and 
\texttt{no\_split} ($E_n$), followed by the M2M counterparts $M_q$, $M_s$ and $M_n$. }
\label{fig:signature-top20-buckets}
\end{figure*}

\begin{figure*}[t]
\centering
\includegraphics[width=\textwidth]{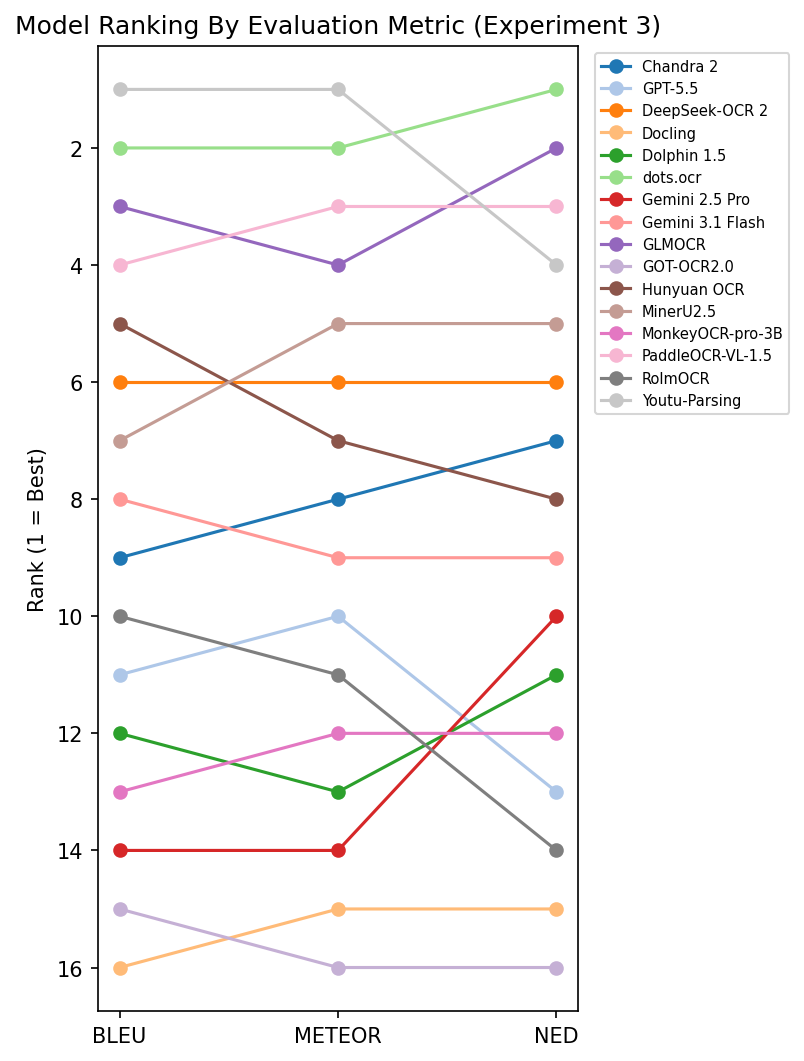}
\caption{Experiment~3 bump chart showing each model's rank trajectory across the three scoring 
metrics: NED, BLEU and METEOR. Rankings are largely preserved, though the top-ranked model 
changes, with dots.ocr leading under NED and Youtu-Parsing under both BLEU and METEOR.}
\label{fig:appendix-bump-metric-exp3}
\end{figure*}

\begin{figure*}[t]
\centering
\includegraphics[width=\textwidth]{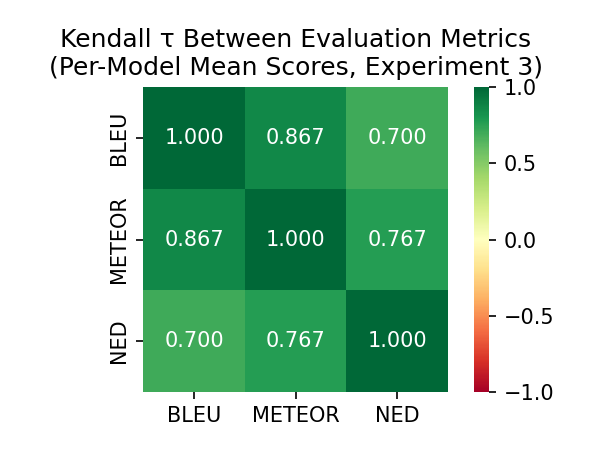}
\caption{Heatmap of pairwise rank agreement across the three Experiment~3 scoring metrics. All 
three pairs show high agreement, with Kendall's $\tau$ between 0.700 and 0.867.}
\label{fig:appendix-heatmap-metric}
\end{figure*}

\clearpage
\section{Detailed Methodology}

This appendix provides additional methodological details and sensitivity analyses supporting the main and supplemental experiments.

\subsection{Implementation Details}
\label{sec:impl}

All experiments were run on NVIDIA H200 GPUs. Inference followed the official OmniDocBench evaluation scripts where available; open-source VLMs were served using vLLM~\cite{kwon2023vllm} where supported.

\textbf{API models.} GPT-5.5, Gemini 2.5 Pro, and Gemini 3.1 Flash were queried using OmniDocBench's reference inference scripts, with the benchmark's fixed system prompt instructing the model to output page content in Markdown. Only the output token cap was set; temperature and sampling parameters were left at provider defaults.

\textbf{Prompt standardisation.} All VLMs received the same page image and a standardised prompt requesting structured markdown output. No model-specific prompt engineering was applied beyond each model's official inference script in OmniDocBench; where no official script existed, the prompt followed the conventions of the closest comparable model.

\renewcommand{\thetable}{C\arabic{table}}
\setcounter{table}{0}
\renewcommand{\thefigure}{C\arabic{figure}}
\setcounter{figure}{0}

\subsection{Alignment Signature Construction}
\label{app:signature-details}

\paragraph{Unit of analysis and scores.}
For every model $m$ and page $p$, we collect the page-level NED score ($1-\text{normalised edit distance}$) under each of the six Experiment~2 alignments. We retain only rows with a valid score under all six alignments ($N=20{,}610$ page$\times$model rows for 16 models).

\paragraph{Discretisation.}
Each score is mapped to one of four buckets:

\begin{center}
\begin{tabular}{clcp{5.5cm}}
\toprule
Bucket & NED range & Label & Interpretation \\
\midrule
\texttt{0} & $\leq 0.01$ & zero & Empty or collapsed match \\
\texttt{L} & $(0.01,\,0.40)$ & low & Poor page-level match \\
\texttt{M} & $[0.40,\,0.85)$ & medium & Partial match \\
\texttt{H} & $\geq 0.85$ & high & Strong match \\
\bottomrule
\end{tabular}
\end{center}

Cutoffs were chosen \emph{a priori} for interpretability rather than from corpus quantiles. Under 
the pooled 16-model distribution they yield bucket fractions \texttt{0}: 2.6\%, \texttt{L}: 8.0\%, 
\texttt{M}: 27.2\%, and \texttt{H}: 62.2\%. Alignment-specific distributions differ markedly: E2E 
\texttt{quick\_match} is heavily right-skewed (median NED $=0.993$; 86.7\% \texttt{H}), whereas M2M 
\texttt{simple\_match} is more dispersed (median $=0.836$; 48.0\% \texttt{H} and 34.2\% \texttt{M}). 
Fixed thresholds therefore encode a common absolute notion of ``good page match'' rather than 
per-alignment relative standing.

We did not rerun the full signature analysis under alternative cutoffs. Qualitatively, the 
conclusions appear robust to modest boundary shifts, since cross-paradigm gaps are large relative 
to bucket width on the affected rows, and the top-20 signatures and E2E-high/M2M-low families 
persist under $\pm 0.05$ perturbations of the \texttt{L}/\texttt{M} boundary in spot checks. Finer 
percentile-based bucketing would increase the observed signature vocabulary without necessarily changing the dominant families.

\paragraph{Signature construction.}
Buckets are concatenated in the fixed order
\begin{center}
$E_q \,\|\, E_s \,\|\, E_n \,\|\, M_q \,\|\, M_s \,\|\, M_n$,
\end{center}
The first three positions denote E2E \texttt{quick\_match}, \texttt{simple\_match}, and \texttt{no\_split}; the final three denote the corresponding M2M methods. For example, \texttt{HHMHHH} denotes high scores under every condition except E2E \texttt{no\_split}, while \texttt{HHHHH0} denotes a collapsed M2M \texttt{no\_split} slot. The vocabulary has at most $4^6=4{,}096$ patterns, of which 301 are observed. Bucketing the NED scores groups rows into recurring types of case, such as those scoring high under every alignment or high under E2E and low under M2M. These groups can be ranked by frequency and inspected as a set, which continuous scores would make impractical since exact repeats are rare. 

\paragraph{Derived structure.}
Each row also receives a \emph{score cluster shape}, describing how raw scores group when rounded to three decimals (e.g.\ \emph{identical}, \emph{E2E $\neq$ M2M}, \emph{E2E split}, \emph{M2M split}, or \emph{loner}), and a \emph{uniformity tier} that distinguishes numerically identical from alignment-sensitive rows. At page level, we compute modal-signature fraction and normalised Shannon entropy. We flag \emph{mechanism-goldmine} pages where at least 65\% of models share one signature, entropy is at most 0.55, and a consensus bad slot (\texttt{0} or \texttt{L}) appears in at least 50\% of model rows.

\paragraph{Failure-mode analysis.}
We proceed in four stages:

\begin{enumerate}
  \item Rank exact signatures by frequency and test model enrichment in each top pattern using one-sided binomial tests with Benjamini--Hochberg FDR correction. The top 20 signatures cover 78.5\% of rows (Figure~\ref{fig:signature-top20-buckets}).
  \item Assign rule-based tags (e.g., \texttt{all\_high} and \texttt{mn\_zero}) and group cross-paradigm families by the E2E--M2M gap direction.
  \item Compare family-level prediction length, format-character ratio, header density, and M2M insertion dominance with corpus baselines and test model-level correlations.
  \item Review up to five distinct pages per \emph{(top-20 signature, cluster-shape)} cell ($\sim490$ exemplars), plus models diverging from the modal signature on approximately 19 mechanism-goldmine pages. Reviewers check for excess or hallucinated content, formatting overhead, region-collapse artefacts, and layout mis-pairing.
\end{enumerate}

\subsection{Diagnostic Protocol Details}
\label{app:diagnostic-details}

The diagnostic reading combines automatic alignment-signature clustering, cross-alignment mechanism summaries, and targeted manual review. Tables~\ref{tab:diagnostic-A} and~\ref{tab:diagnostic-B} summarise frequent recurring families on OmniDocBench v1.5. Frequencies are the share of listed exemplar signatures in each row; the rows are illustrative families rather than an exhaustive partition.

E2E-favouring signatures listed in Table~\ref{tab:diagnostic-A} account for approximately 25\% of page$\times$model rows (12\% excess-content and 13\% format-overhead families), compared with approximately 0.8\% for the listed M2M-favouring signatures. The broader pattern ``E2E all-high, M2M not all-high'' covers approximately 17\% of rows, whereas M2M-favouring rows with all-E2E \texttt{0}/\texttt{L} slots account for approximately 0.2\%.

This frequency gap does not establish that E2E is globally superior. OmniDocBench is dominated by cases where region-based matching is lenient on extra or loosely aligned text and full-page Markdown comparison is stricter. Math-heavy benchmarks may produce more M2M$>$E2E equation-mispairing cases such as \texttt{LLLHHH} or \texttt{LLMHMM}; title-heavy slides or worksheets may produce more \texttt{00*HHH}-type region-collapse cases.

\begin{table}[H]
\centering
\footnotesize
\setlength{\tabcolsep}{3pt}
\caption{Diagnostic families where E2E $>$ M2M (OmniDocBench v1.5, 16 models; $N=20{,}610$ page$\times$model rows). Frequencies are the combined share of listed signatures.}
\label{tab:diagnostic-A}
\resizebox{\textwidth}{!}{%
\begin{tabular}{>{\raggedright\arraybackslash}p{2.2cm} >{\raggedright\arraybackslash}p{2.4cm} >{\raggedright\arraybackslash}p{2.0cm} >{\raggedright\arraybackslash}p{3.2cm} r >{\raggedright\arraybackslash}p{2.4cm}}
\toprule
\textbf{Observed pattern} & \textbf{Likely cause} & \textbf{What to check} & \textbf{Example signatures} & \textbf{Freq.} & \textbf{Recommendation} \\
\midrule
E2E high, M2M lower & Excess content (hallucinations, captions, duplication) & Pred.\ length vs.\ GT; extra spans & \texttt{HHMMMM}, \texttt{HHHMMM}, \texttt{HHMHMM}, \texttt{HMHHMM}, \texttt{HMMMMM}, \texttt{HHLLLL}, \texttt{HHMMML} & $\sim12\%$ & Report both protocols; inspect extra output when E2E $\gg$ M2M \\
\addlinespace
E2E high, M2M low/medium, heavy syntax & Formatting overhead (Markdown/HTML penalised byte-wise) & Headers/HTML; stripped vs.\ full Markdown & \texttt{HHHHHM}, \texttt{HHHHMH}, \texttt{HMHHLM}, \texttt{HHHHLH}, \texttt{HLHHLH}, \texttt{HHMHLM} & $\sim13\%$ & Pair region-based scores with M2M; report syntax-stripped variants \\
\bottomrule
\end{tabular}%
}
\end{table}

\begin{table}[H]
\centering
\footnotesize
\setlength{\tabcolsep}{3pt}
\caption{Diagnostic families where M2M $>$ E2E (same corpus; rare on OmniDocBench v1.5).}
\label{tab:diagnostic-B}
\resizebox{\textwidth}{!}{%
\begin{tabular}{>{\raggedright\arraybackslash}p{2.2cm} >{\raggedright\arraybackslash}p{2.4cm} >{\raggedright\arraybackslash}p{2.0cm} >{\raggedright\arraybackslash}p{3.2cm} r >{\raggedright\arraybackslash}p{2.4cm}}
\toprule
\textbf{Observed pattern} & \textbf{Likely cause} & \textbf{What to check} & \textbf{Example signatures} & \textbf{Freq.} & \textbf{Recommendation} \\
\midrule
E2E low, M2M high (math) & Equation/block mis-pair under E2E & Equation blocks in GT & \texttt{LLLHHH}, \texttt{LLLHMM}, \texttt{LLMHMM} & $\sim0.02\%$ & Stratify math-heavy pages; audit block matching \\
\addlinespace
E2E low, M2M high (layout) & Column/reading-order break under concatenation & Block order vs.\ M2M concatenation & \texttt{MMMHHH}, \texttt{MMMHMH}, \texttt{MMLHMH}, \texttt{HLMHLH}, \texttt{LLLMML}, \texttt{00LLLL} & $\sim0.8\%$ & Report M2M on layout-heavy pages \\
\addlinespace
E2E title/box slot empty, M2M body correct & Region-slot failure (title dropped) & GT title region vs.\ E2E match & \texttt{00MHHH}, \texttt{00HHHH}, \texttt{00MHHM} & $\sim0.03\%$ & Inspect region-level matching, not page OCR quality \\
\bottomrule
\end{tabular}%
}
\end{table}

\FloatBarrier
\clearpage

\end{document}